\documentclass[letterpaper,journal]{IEEEtran}
\usepackage{amsmath,amsfonts}
\usepackage{lineno}
\usepackage{algorithm}
\usepackage{array}
\usepackage[caption=false,font=normalsize,labelfont=sf,textfont=sf]{subfig}
\usepackage{textcomp}
\usepackage{stfloats}
\usepackage{url}
\usepackage{verbatim}
\usepackage{graphicx}
\usepackage{cite}
\usepackage{algorithm,algorithmicx,algpseudocode }
\usepackage{multirow}
\usepackage{subfig}
\usepackage{graphicx}
\usepackage[utf8]{inputenc}
\usepackage{diagbox}
\usepackage{xcolor}
\usepackage{booktabs}

\begin{document}

\title{Flow of Truth: Proactive Temporal Forensics for Image-to-Video Generation}

\author{
Yuzhuo Chen, 
Zehua Ma*,
Han Fang*,
Hengyi Wang,
Guanjie Wang,
Weiming Zhang
\thanks{This work was supported in part by the Natural Science Foundation of China under Grant 62402469, 62121002, 62472398, U2336206.

Yuzhuo Chen, 
Zehua Ma,
Han Fang,
Hengyi Wang,
Guanjie Wang,
Weiming Zhang are with the Anhui Province Key Laboratory of Digital Security (School of Cyber Science and Technology, University of Science and Technology of China), Hefei 230026, China.

*Corresponding author: Zehua Ma (Email: mzh045@ustc.edu.cn), Han Fang (Email: fanghan@ustc.edu.cn)}
}



\maketitle

\begin{abstract}

The rapid rise of image-to-video (I2V) generation enables realistic videos to be created from a single image but also brings new forensic demands. Unlike static images, I2V content evolves over time, requiring proactive forensics to move beyond spatial tampering localization toward tracing how protected evidence flows and transforms throughout the video. As frames progress, embedded traces may drift and deform, making static spatial forensics unreliable in this setting. To address this unexplored problem, we present \textbf{Flow of Truth}, a proactive framework for temporal traceability in I2V generation. A key challenge is discovering a forensic signature that can remain synchronized with a generative process that may introduce motion, occlusion, and semantic re-synthesis. We therefore model I2V generation from a pixel-motion perspective. Building on this view, we propose a learnable forensic template that follows pixel motion and a template-guided flow prediction module that decouples motion from image content, enabling source recovery and temporal tracing. Experiments across commercial and open-source I2V models show that Flow of Truth provides an effective first step toward proactive temporal forensics.

\end{abstract}

\section{Introduction}
\label{sec:intro}

The rapid advancement of image-to-video (I2V) generation models, such as AnimateDiff \cite{videogen-AnimateDiff}, Wan2.2 \cite{videogen-wan}, Sora2 \cite{videogen-sora}, Veo3 \cite{videogen-veo}, and Kling \cite{videogen-kling}, has made it possible to synthesize highly realistic and temporally coherent videos from a single image, boosting
creative production, industrial content generation, and automated media workflows. However, they also introduce new risks of misinformation and large-scale content fabrication, highlighting an urgent need for more reliable forensic mechanisms capable of verifying the authenticity of automatically generated videos.


Traditional image editing forensics focuses on identifying which spatial regions of an image have been manipulated. Most proactive forensic systems \cite{wm_and_tl-Editguard, wm_and_tl-OmniGuard} typically embed a forensic template into the static image and later recover it to localize tampered areas. However, I2V generation creates a different forensic problem. Instead of performing isolated edits, I2V models produce videos through progressive visual evolution, where pixels may move, deform, disappear, or be semantically re-synthesized as they propagate through time. Thus, identifying where manipulation occurs in one frame is no longer sufficient; a proactive I2V forensic system must also trace how protected evidence moves across time and recover the source image behind the generated video. This temporal drift causes embedded forensic traces to misalign or dissipate, making static spatial forensics unreliable in the I2V setting.


Achieving temporal forensics in the I2V setting is intrinsically challenging. The core difficulty lies in discovering a forensic signal that can co-evolve with the generative process. Unlike deterministic image edits confined to the original pixel layout, I2V generation continuously synthesizes new content and reshapes pixel trajectories, leading both appearance and semantics to drift unpredictably. Embedded forensic evidence can easily be overwritten or desynchronized, making it nontrivial to maintain a stable, traceable representation throughout the video.

To address this challenge, we introduce Flow of Truth (FoT), the first proactive forensic framework for source recovery and temporal traceability in I2V generation. The central obstacle is that, after misuse, the defender typically observes only the video frames released by the attacker; the \textbf{original source image is no longer available} as an input for standard \textbf{source-target motion estimation}. The central idea is therefore to interpret I2V generation from a pixel-motion perspective rather than only as frame synthesis. Under this perspective, temporal forensics becomes tractable: if a forensic signal is embedded in the source before generation and is designed to ``follow'' and ``translate'' underlying pixel dynamics, it can remain synchronized with the evolving target frames.


Building on this motion-based formulation, we operationalize temporal forensics by embedding a learnable forensic template into the protected image. The template acts as a \textbf{source-side anchor} that is carried into the target video together with the image content. During inference, FoT decodes the surviving template evidence from each released target frame and uses it as an image-independent cue for source-to-frame correspondence, avoiding the need to observe the source frame alongside the target frame. To align this correspondence with real video dynamics, we introduce a \textbf{template-guided flow prediction} module and jointly optimize it with the forensic template, ensuring that the embedded signal remains synchronized with pixel motion and consistently traceable throughout the generated video.

Extensive experiments show that FoT remains effective across both commercial and open-source I2V models. Because existing image forensics, watermarking, and optical-flow methods either assume static spatial alignment or require \textbf{unavailable source-target image pairs}, they do not directly provide source recovery from I2V frames with a proactive template. Accordingly, we evaluate FoT with task-specific references and auxiliary forensic tasks. The main contributions are as follows:



\begin{itemize}
    \item We identify proactive temporal traceability as a new forensic requirement for I2V generation, where protected evidence must be traced through motion and semantic changes.
    \item We propose \textbf{Flow of Truth (FoT)}, a new proactive forensic framework that reframes I2V generation from a pixel-flow perspective. By modeling video generation as pixel-wise motion rather than frame synthesis, FoT learns a forensic template that co-evolves with pixel trajectories, enabling a faithful characterization of both the motion dynamics and the generative transformation process.
    \item We conduct extensive evaluations across both open-source and commercial I2V models, demonstrating strong generalization and high temporal traceability.
    \item We further demonstrate that FoT acts as a general plug-in module, improving the robustness of copyright watermarking against geometric distortions and boosting resistance to I2V transformations for compression-robust watermarks. Furthermore, FoT-embedded images show inherent resilience to pixel-aligned tampering, suggesting a possible pathway toward a more broadly protective forensic information carrier.

\end{itemize}


\begin{figure*}[htbp]
    \centering
    \includegraphics[width=0.95\linewidth]{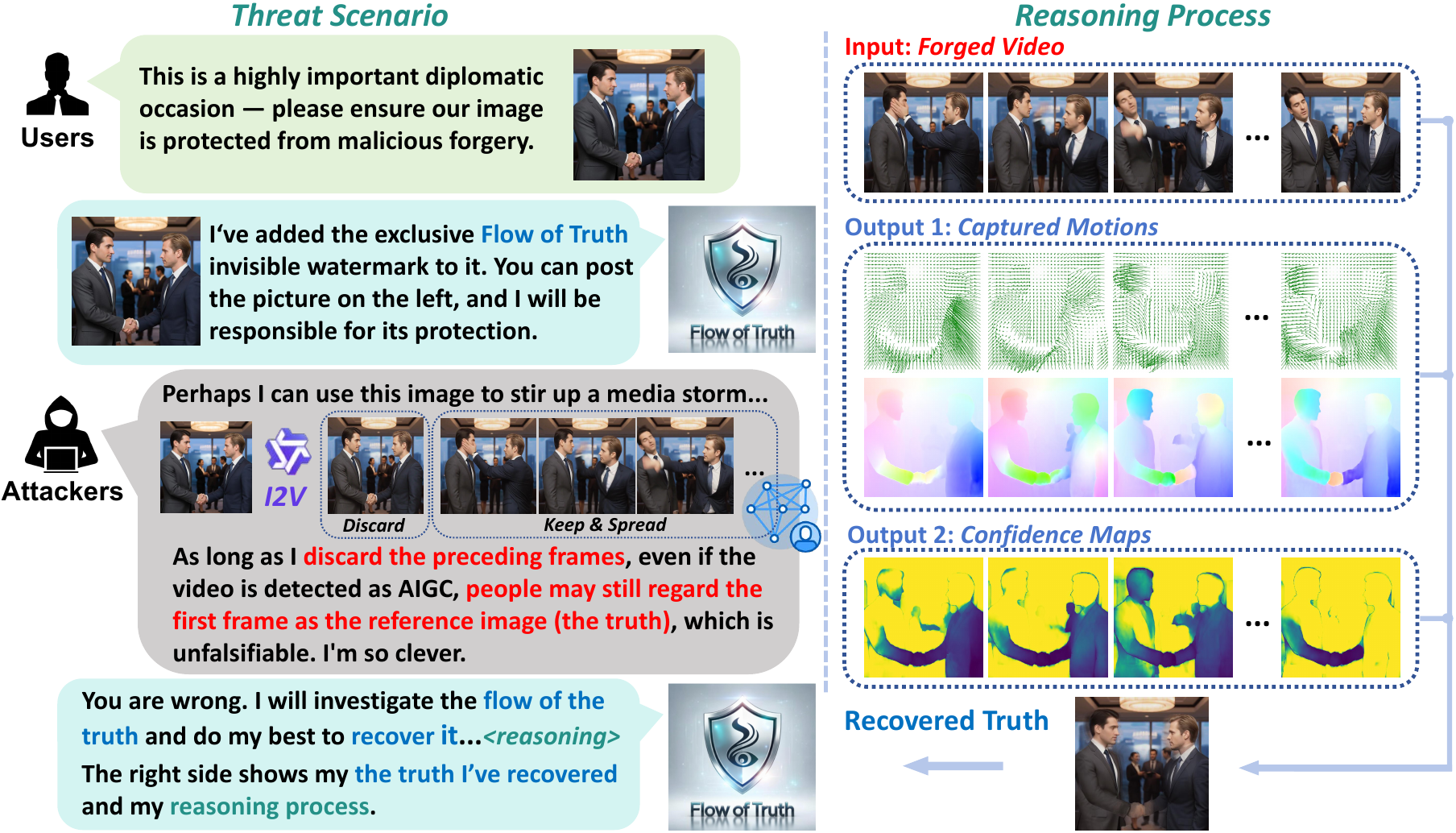}
    \caption{Overview of the proactive temporal forensic setting addressed by \textbf{Flow of Truth (FoT)}. \textbf{Left:} a source image is protected by embedding a learnable forensic template and is then transformed by an I2V generator into a forged video, where the embedded evidence may drift, deform, or become partially occluded across frames. \textbf{Right:} FoT decodes the surviving template evidence from the generated frames, predicts source-to-frame motion, and reverses the frames back to the source coordinate system, enabling recovery of the truthful source behind the I2V manipulation.}
    \label{fig:teaser}
\end{figure*}

\section{Related Works}

\noindent\textbf{Proactive Image Forensics.}
Proactive forensics embeds verifiable signals to support authenticity verification~\cite{postprocessing_watermarking-REVMark, postprocessing_watermarking-RoPaSS} and tampering localization~\cite{wm_and_tl-Editguard, wm_and_tl-OmniGuard}.
Traditional watermarking or residual-based schemes offer robustness to mild edits but fail under semantic or structural changes.
Recent encoder–decoder based forensic embedding improves fidelity–robustness trade-offs, yet these methods inherently assume spatial alignment and operate only on static images, making the embedded traces unstable when the image is transformed into a dynamic video.
FoT is complementary to these methods: it protects a source image before potential I2V misuse and aims to recover temporal evidence after the protected image has been transformed into video, rather than only verifying a static image or localizing a static edit.

\noindent\textbf{Image-to-Video (I2V) Generation.}
Modern I2V systems—including Wan 2.2~\cite{videogen-wan}, MovieGen~\cite{videogen-MovieGen}, and Hunyuan Video~\cite{videogen-Hunyuan}—use diffusion or transformer architectures to produce temporally coherent videos from a single image.
Closed-source models such as Sora~\cite{videogen-sora}, Veo 3~\cite{videogen-veo}, and Kling~\cite{videogen-kling} further improve realism but prevent model-specific forensic integration and large-scale analysis.
Embedding full I2V generators into forensic training is computationally prohibitive, and for closed-source systems it is not directly possible. Even for open-source generators, such training may introduce model-specific bias because the forensic signal can adapt to one architecture rather than to transferable temporal evidence.

\noindent\textbf{Optical Flow and Motion Representation.}
Optical flow networks (e.g., RAFT~\cite{ofmodel-RAFT}, GMFlow~\cite{ofmodel-GMFlow}, FlowFormer++~\cite{ofmodel-FlowFormer++}) estimate pixel-wise motion through correlation or transformer-based reasoning, but they normally require both \textbf{source and target RGB frames} as input.
This input assumption does not hold in our forensic setting: after an attacker releases an I2V-generated video, the defender can observe only the target video frames, while the \textbf{original source image is unavailable} in the evidence stream.
This motivates \textbf{\textit{template-guided flow evolution}}: FoT embeds a learnable template into the protected source image, lets the template evolve with the image content during I2V generation, and then extracts the surviving template evidence from each target frame to infer source-to-frame correspondence for recovery.

\begin{figure*}[htbp]
    \centering
    \includegraphics[width=\linewidth]{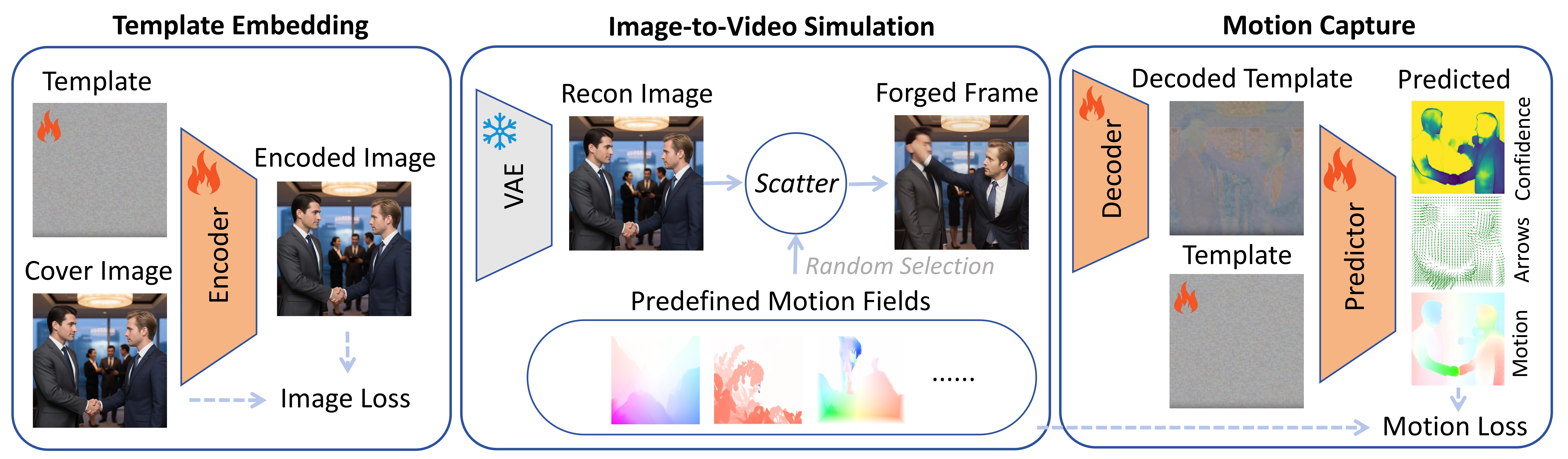}
    \caption{
        Training of \textbf{Flow of Truth (FoT)}. 
        (1) \textbf{Template Embedding}, which implants a learnable forensic template into an image while preserving its fidelity;
        (2) \textbf{Image-to-Video Simulation}, which emulates compression and motion-induced evidence drift without inserting a full I2V generator into the training loop; 
        (3) \textbf{Motion Capture}, which predicts template-guided source-to-frame correspondence for later flow reversal.
    }
    \label{fig:training_pipeline}
\end{figure*}

\section{Method}

\subsection{Task Reformulation for I2V Forensics}
\noindent
Conventional image forensics methods typically assume static pixel alignment between the original and manipulated contents. 
However, in image-to-video (I2V) generation, a single image is transformed into a sequence of frames through dynamic motion synthesis, 
causing spatial forensic traces to drift over time. 
We therefore reformulate I2V forensics as a \textit{temporal correspondence problem}: the objective is not merely to detect tampered pixels in a single frame, but to trace how forensic information propagates across frames and ultimately recover the original image.

Given a source image $I_0 \in \mathbb{R}^{3 \times H \times W}$ and its generated video frames 
$\{I_t\}_{t=1}^{N}$, the goal is to learn a mapping
\begin{equation}
    \Phi: I_t \mapsto \widetilde{I}_{t \rightarrow 0}
\label{eq:mapping t to 0}
\end{equation}
where $\widetilde{I}_{t \rightarrow 0}$ denotes the recovered image from frame $t$, 
which approximates the original source $I_0$. 
In practice, $\Phi$ receives only an observed frame $I_t$ and reconstructs its motion-aligned counterpart $\widetilde{I}_{t\rightarrow 0}$. 
The final reconstruction is obtained by aggregating all frame-wise predictions:
\begin{equation}
    \widetilde{I}_0 = \Psi(\{\widetilde{I}_{t\rightarrow 0}\}_{t=k}^{N}),k\geq 1
\label{eq:aggregate}
\end{equation}
where $\Psi(\cdot)$ is a fusion operator that integrates multi-frame evidence to restore a temporally consistent and forensic-aligned image. Here $k$ is the index of the first available frame: $k=1$ means that no initial frame is removed, while $k>1$ represents an attacker discarding the first $k-1$ frames before forensic analysis.

\subsection{Overview of FoT}
\noindent
To learn the mapping $\Phi$ defined in Eq.~(\ref{eq:mapping t to 0}), 
a straightforward idea is to collect paired data $(I_t, \widetilde{I}_{t\rightarrow 0})$ 
and train an end-to-end neural network. 
However, this setup is inherently \textbf{ill-posed}:
given only $I_t$, infinitely many visually plausible reconstructions $\widetilde{I}_{t\rightarrow 0}$ exist. 
Thus, embedding a forensic trace to constrain this mapping is essential (Sec.~\ref{sec:template embedding}).

Then, how can we obtain such paired data in a differentiable way?  
Embedding an I2V model directly into the training pipeline is theoretically feasible but highly inefficient, unavailable for closed-source generators, and prone to model-specific bias. Hence, we use a differentiable simulation that captures transferable compression and motion factors instead of training against one generator (Sec.~\ref{sec:I2V simulation}).  

Furthermore, we reformulate the generative problem into a more tractable discriminative one: 
learning a dense displacement field from $I_0$ to $I_t$. 
This reformulation not only enables pixel-level interpretability of motion 
but also allows us to use the learned field to backward-warp $I_t$ into $\widetilde{I}_{t\rightarrow 0}$.

In summary, motivated by these insights, 
we design a four-stage pipeline to ensure that forensic evidence remains traceable across motion:  
\textbf{1) Template Embedding}, which implants forensic cues into the source image;  
\textbf{2) I2V Simulation}, which exposes the embedded information to motion transformations;  
\textbf{3) Motion Capture}, which learns the motion correspondence between images; and  
\textbf{4) Flow Reversal}, which recovers source images during inference.

The first three stages constitute the differentiable training process, 
while the last stage operates during inference to reconstruct the final forensic-consistent image.
These stages are deliberately coupled: the template supplies the forensic carrier, simulation provides differentiable motion labels, motion capture converts template evidence into source-to-frame correspondence, and flow reversal turns that correspondence back into source evidence. Removing any stage would change the task interface rather than merely disable an optional refinement.

\subsection{Template Embedding}
\label{sec:template embedding}
\noindent
In the task of tampering localization, EditGuard \cite{wm_and_tl-Editguard} embeds a pure blue image into the cover as a fragile watermark to localize tampered regions. OmniGuard \cite{wm_and_tl-OmniGuard} explored more diverse template types and found that using fixed natural images, coupled with reformulating the reconstruction task as a classification problem, can achieve higher visual fidelity. However, both methods rely on fixed templates. 

While fixed templates are sufficient for localization tasks, they are insufficient for our setting. Here, the template must be complex enough to track pixel-level motion, yet refined enough to preserve visual fidelity.

To address this, we adopt a \textit{learnable forensic template}, treating it as part of the model's internal representation and supervising it through comprehensive, task-level objective functions. Specifically, we define an encoder $\mathcal{E}$ and a learnable template $T \in \mathbb{R}^{C \times H \times W}$. The encoder integrates the template into the cover image $I_0 \in \mathbb{R}^{3 \times H \times W}$:
\begin{equation}
    I_0^T = \mathcal{E}(I_0, T)
\end{equation}
where $I_0^T$ visually resembles $I_0$ while carrying imperceptible template information. 
Unlike conventional watermarking, FoT encourages \textit{motion-aligned embedding}, where the encoder allocates template energy to visually stable regions likely to maintain temporal consistency during image-to-video synthesis. 
Following learned data hiding and watermarking systems~\cite{postprocessing_watermarking-HiDDeN,postprocessing_watermarking-Stegastamp}, a reconstruction loss combines pixel distortion with the perceptual distance LPIPS~\cite{metrics-LPIPS} to keep the embedded template visually unobtrusive:
\begin{equation}
    \mathcal{L}_{\text{img}} = \lambda_1 \| I_0 - I_0^T \|_2^2 + \lambda_2 \text{LPIPS}(I_0, I_0^T)
\end{equation}

The template and motion alignment are jointly optimized via dedicated loss functions, which are detailed in Sec.~\ref{sec:motion capture}.

\subsection{Image-to-Video Simulation}
\label{sec:I2V simulation}
\noindent
To mimic the temporal deformation caused by I2V generators without inserting real I2V models into training, we introduce a differentiable simulation mechanism. Specifically, we abstract I2V generation into two transferable sub-processes: \textit{(1) feature compression and reconstruction, and (2) feature flow.}

For the first sub-process, we employ a frozen variational autoencoder (VAE) to approximate the compression-reconstruction dynamics of I2V models. 
To avoid overfitting the specific VAE's latent space, we reconstruct the image directly after compression to obtain $\hat{I_0^T} = \text{VAE}(I_0^T)$.

Next, to emulate temporal dynamics, we scatter the pixels of $\hat{I_0^T}$ using a motion field $M_i$ randomly sampled from a predefined bank $\mathcal{M}$ that captures plausible video motions:
\begin{equation}
    I^{T_t}_t = \mathcal{S}(\hat{I_0^T}, M_i), \quad M_i \sim \mathcal{M}
\end{equation}
where $\mathcal{S}$ denotes a differentiable scatter operator that warps pixels according to $M_i$. 
The resulting ``forged image'' $I^{T_t}_t$ simulates a potential I2V frame under motion $M_i\in \mathbb{R}^{2\times H\times W}$, accompanied by a motion-aware template $T_f$.
The bank $\mathcal{M}$ is built from optical-flow datasets used in our training set: FlyingChairs provides large-scale planar synthetic object/background motion, FlyingChairs2 adds occlusions and motion boundaries, Sintel introduces long-range and non-rigid movie-like motion, and Spring contributes high-resolution detailed scene motion. This bank is not intended to reproduce every I2V semantic transformation. Instead, it supplies controllable supervision for learning how the template should move with pixels.

Here, we assume that through learning, each spatial feature vector can evolve with a motion pattern consistent with its corresponding pixel in the image. This assumption serves as a prerequisite for subsequent motion capture.

By incorporating the VAE, FoT regularizes the embedding space to remain stable under compression and reconstruction, enhancing robustness against appearance changes.  
By stochastically sampling motion fields from $\mathcal{M}$, FoT further learns to resist diverse pixel displacements, simulating the temporal distortions observed in real I2V generation.  
\textbf{This design also provides explicit ground-truth supervision for training: each synthetic motion field $M_i$ serves as a known label that directly supervises the learning of the motion-aligned forensic template.}  
This unique property allows the main differentiable training of FoT to avoid large-scale real I2V videos and to remain independent of any specific I2V generator, while still capturing transferable temporal behavior.

\subsection{Motion Capture}
\label{sec:motion capture}
This stage aims to estimate the dense displacement field $F_{0\rightarrow t}$ between the embedded source image $I_0^T$ and its simulated I2V frame $I_t^{T_t}$, where $I_0^T$ is unavailable.

\noindent\textbf{Template Decoding.}
We first decode the motion-aware template $T_t = \mathcal{D}(I_t^{T_t})$ using a trainable decoder $\mathcal{D}(\cdot)$.

\noindent\textbf{Motion Estimation Under Uncertainty.}
Unlike conventional restoration tasks, forensic recovery must cope with regions that are unreliable or inherently unobservable—e.g., pixels visible in $I_0$ but missing in later frames due to occlusion or generative randomness. 
To handle this, we integrate \textit{uncertainty prediction} into the motion estimation process, enabling the model to explicitly quantify its confidence in each motion vector.
Following prior probabilistic formulations~\cite{ofmodel-SEA-RAFT}, we adopt the Mixture-of-Laplace loss:
\vspace{-0.2cm}
\begin{equation}
\scalebox{1.03}{$
    \mathcal{L}_{\text{MoL}}
    = -\log[\alpha \cdot \frac{e^{-\frac{|\mathbf{v}_\text{p}-\mathbf{v}_\text{gt}|}{e^{\beta_1}}}}{2e^{\beta_1}} + (1-\alpha) \cdot \frac{e^{-\frac{|\mathbf{v}_\text{p}-\mathbf{v}_\text{gt}|}{e^{\beta_2}}}}{2e^{\beta_2}}]
    $}
\end{equation}
with a well-designed optical flow estimator $\mathcal{P_o}(\cdot)$ to predict the flow and required parameters:
\begin{equation}
    F_{0\rightarrow t},\alpha,\beta_1,\beta_2 = \mathcal{P_o}(T, T_t)
\end{equation}
where $\mathbf{v_p}\in F_{0\rightarrow t}$ and $\mathbf{v_{gt}}\in M_{i}$ respectively denote predicted and ground-truth motion vectors, and $\alpha$, $\beta_1$, and $\beta_2$ are predicted parameters. 
Here, $\beta_1$ is fixed to $0$ throughout training and inference so that the first component represents deterministic motion, while the second models uncertain cases such as occlusion, semantic re-synthesis, or newly generated content.

Given that a higher $\alpha$ indicates greater confidence, and a smaller Laplace scale $b$ (e.g., $e^{\beta_1}$ or $e^{\beta_2}$) implies a more concentrated, i.e., more deterministic, distribution, we define a normalized confidence map:
\begin{equation}
    \mathcal{C}_{\text{map}} = \text{Norm}\!\left(\frac{\alpha}{2e^{\beta_1}} + \frac{1-\alpha}{2e^{\beta_2}}\right)
\end{equation}
which down-weights unreliable flow vectors, especially in occluded or motion-ambiguous regions, thereby improving robustness in multi-frame forensic reconstruction.

The total training objective is finally defined as:
\begin{equation}
    \mathcal{L}_{\text{total}} = \mathcal{L}_{\text{img}} + \lambda_m \mathcal{L}_{\text{MoL}}
\end{equation}
where $\mathcal{L}_{\text{img}}$ enforces image fidelity, while $\mathcal{L}_{\text{MoL}}$ achieves motion capture by explicitly modeling uncertainty.

\subsection{Flow Reversal}
\label{sec:flow reversal}
\noindent
During inference, FoT employs the predicted motion fields to align each frame back to the source image 
and aggregates the recovered results using confidence-guided fusion.

\paragraph{Implementation of $\Phi$ in Eq.~(\ref{eq:mapping t to 0})}
Given the predicted flow $F_{0\rightarrow t}$, the motion-aligned reconstruction is obtained via backward warping:

\begin{equation}
\scalebox{0.9}{$
    \widetilde{I}_{t\rightarrow0} = \mathcal{W}(I_t, F_{0\rightarrow t}) = I_t(x + F_{0\rightarrow t}(x), y + F_{0\rightarrow t}(y))
$}
\end{equation}
where $\mathcal{W}(\cdot)$ is a differentiable bilinear warping operator, and $(x, y)$ are spatial coordinates.

\paragraph{Implementation of $\Psi$ in Eq.~(\ref{eq:aggregate})}
After obtaining all motion-reversed frames, FoT fuses them to produce the final recovered image using confidence maps as soft masks:

\begin{equation}
    \widetilde{I}_0 = \frac{\sum_{t=k}^{N}\mathcal{C}_{\text{map},0\rightarrow t} \odot \widetilde{I}_{t\rightarrow0}}{\sum_{t=k}^{N}\mathcal{C}_{\text{map},0\rightarrow t}}, \quad k\ge1.
\end{equation}
In summary, 
FoT not only visually restores the truth but ensures that forensic evidence remains temporally traceable throughout the entire I2V process.

\section{Experiments}
\label{Experiments}

\subsection{Implementation Details}

\noindent\textbf{Data Curation.}
Our training set combines 118K MSCOCO~\cite{datasets-MSCOCO} images and 85K optical flow samples collected from FlyingChairs~\cite{datasets-of-FlyingChiars}, FlyingChairs2~\cite{datasets-of-FlyingChiars2}, Sintel~\cite{datasets-of-Sintel}, and Spring~\cite{datasets-of-Spring}. These flow datasets provide complementary supervision: planar synthetic motion, occlusions and motion boundaries, long-range non-rigid movie motion, and high-resolution detailed scene motion. All data are resized to $512{\times}512$, which exposes the embedder and decoder to resolution normalization during training.  
For evaluation, we construct a forensic-oriented benchmark covering five motion scenarios: \textit{Face}, \textit{Camera}, \textit{Animal}, \textit{Human-Environment (H-E)}, and \textit{Multi-Human (M-H)}.  
As detailed in Table~\ref{tab:curated evaluation dataset}, the benchmark contains 204 source-prompt pairs: 50 FFHQ-1024~\cite{datasets-FFHQ} face images, 35 DIV2K~\cite{datasets-DIV2K} landscapes for camera motion, 19 DIV2K animal images, 22 single-person MSCOCO images with both subject-only and subject-camera prompts, and 28 multi-person MSCOCO images with subject and subject-camera prompts.
The camera-motion subset covers translational movements (Dolly In/Out, Truck, and Pedestal), rotational movements (Pan, Tilt, and Roll), lens operations (Zoom In/Out), and challenging handheld motion.
Each scenario is paired with manually crafted prompts that keep the requested motion physically plausible and consistent with scene semantics. For example, face and animal prompts emphasize local subject motion under a fixed camera, whereas camera prompts explicitly keep scene objects still and specify the camera trajectory.
Videos are generated using four representative I2V models: CogVideoX-5B-I2V~\cite{videogen-CogVideoX} and Wan2.2-I2V-A14B~\cite{videogen-wan}, which produce 49-frame videos, and Kling2.1~\cite{videogen-kling} and Dreamina S2.0~\cite{videogen-dreamina}, which produce 121-frame videos. This yields 816 videos and 69,360 frames in total for the main experiments. Unlike standard deepfake benchmarks that mainly support real/fake classification, this benchmark keeps the protected source image, prompt category, generated video, and motion type together so that source recovery and temporal traceability can be evaluated.

\noindent\textbf{Prompt Templates.}
To make the benchmark reproducible and to avoid mixing semantically incompatible motions, each prompt is decomposed into source-scene description, subject/camera motion instruction, and camera-state constraint. 
Table~\ref{tab:prompt_templates} provides one representative template for each scenario type, illustrating how source-scene description, motion instruction, and camera constraint are combined.
\begin{table*}[!t]
\centering
\caption{Representative prompt templates used in the evaluation benchmark. Each prompt specifies the source scene, requested motion, and camera constraint.}
\label{tab:prompt_templates}
\footnotesize
\setlength{\tabcolsep}{0.045in}
\begin{tabular}{p{0.12\textwidth}p{0.82\textwidth}}
\toprule
\textbf{Scenario} & \textbf{Representative Template} \\ \midrule
\textit{Face} &
\textit{A woman smiling with her hands under her chin. The expression gradually becomes serious. Camera stays perfectly still.} \\
\textit{Camera} &
\textit{A symmetrical artwork with radiating lines. The objects in the picture remain still. The camera slowly rolls and zooms out.} \\
\textit{Animal} &
\textit{A vibrant red bird perches on a branch with red flowers. It ruffles its feathers slightly and then settles. Camera stays perfectly still.} \\
\textit{H-E} &
\textit{A man takes baked food out of the oven. He uses too much force, causing the food in his left hand to hit his face. Camera remains still.} \\
\textit{M-H} &
\textit{Three people are playing football while others watch. The person on the far left suddenly tackles the ball holder. The camera slowly dollies in.} \\
\bottomrule
\end{tabular}
\end{table*}

\begin{table}[!t]
\centering
\caption{Details of the curated evaluation dataset. The values represent the number of source images or prompts for each I2V motion type.}
\label{tab:curated evaluation dataset}
\resizebox{0.94\linewidth}{!}{
\begin{tabular}{lccccc}
\toprule
\textbf{Type} & \textbf{Face} & \textbf{Camera} & \textbf{Animal} & \textbf{H-E} & \textbf{M-H} \\ \midrule
Subject Motion & 50 & 0 & 19 & 22 & 28 \\
Camera Motion & 0 & 35 & 0 & 0 & 0 \\
Subject \& Camera Motion & 0 & 0 & 0 & 22 & 28 \\ \midrule
Total & 50 & 35 & 19 & 44 & 56 \\ \bottomrule
\end{tabular}
}
\end{table}

\noindent\textbf{Architecture.}
FoT uses a UNet backbone with windowed self-attention, and the channel size of the learnable forensic template $T$ is set to 3, to align with a flow predictor built upon SEA-RAFT pre-trained on large-scale datasets.
This design stabilizes training and ensures reliable template-guided motion modeling.  

\noindent\textbf{Training Implementation.}
We implement the framework in PyTorch and train it on four NVIDIA RTX 3090 GPUs using AdamW.
Training proceeds in three stages:
1) joint optimization of all modules for 190K iterations with $\lambda_1{=}\lambda_2{=}1.0$, $\lambda_m{=}0.1$, and learning rate $\eta{=}1{\times}10^{-4}$;
2) encoder refinement for 88K iterations with frozen decoder/predictor, $\lambda_1{=}10$, and $\eta{=}5{\times}10^{-5}$;
3) final fine-tuning of decoder and predictor for 4K iterations on 20 Kling2.1-generated videos, with frozen encoder/template, $\lambda_1{=}\lambda_2{=}0$, $\lambda_m{=}1.0$, and $\eta{=}5{\times}10^{-5}$. The last stage utilizes a small batch of high-quality generated samples to efficiently facilitate the distribution-adaptation of the decoder and predictor to real generated videos.

\noindent\textbf{Image Fidelity of Embedded Sources.}
FoT maintains high perceptual quality throughout all experiments: the encoded images reach PSNR~36.58 / SSIM~0.9627 / LPIPS~0.0250 on COCO2017-val (5K samples), and 36.33 / 0.9655 / 0.0236 on DIV2K (100 samples). These results are reported here as a prerequisite for the recovery experiments because proactive forensics must preserve source-image quality before any I2V misuse occurs.

\noindent\textbf{Degradation Protocol.}
For robustness evaluation, each degraded input is processed by one randomly selected transformation: JPEG compression with quality in $[30,90]$, Gaussian noise with zero mean and standard deviation in $[1,5]$, Gaussian blur with radius $1$ or $2$, median blur with kernel size in $\{3,5,7,9\}$, resize-and-recover with ratio in $[0.6,0.9]$, or brightness adjustment with factor $1$ or $2$.
Table~\ref{tab:degradation_protocol} summarizes the exact degradation pool.
Only one degradation is applied to each sample, so the robustness measurement isolates the effect of individual post-processing operations instead of conflating several distortions.
This design matches practical forensic use cases where generated frames are often affected by one dominant operation, such as platform recompression, resizing, mild blur, or illumination adjustment, before analysis.
\begin{table}[t]
\centering
\caption{Random image degradation protocol used for robustness evaluation. One degradation type and one strength are sampled for each degraded test input.}
\label{tab:degradation_protocol}
\setlength{\tabcolsep}{0.035in}
\begin{tabular}{p{0.34\linewidth}p{0.56\linewidth}}
\toprule
\textbf{Degradation} & \textbf{Random Strength} \\ \midrule
JPEG compression & quality $\in[30,90]$ \\
Gaussian noise & mean $=0$, standard deviation $\in[1,5]$ \\
Gaussian blur & radius $\in\{1,2\}$ \\
Median blur & kernel size $\in\{3,5,7,9\}$ \\
Resize and recover & ratio $\in[0.6,0.9]$ \\
Brightness transform & factor $\in\{1,2\}$ \\
\bottomrule
\end{tabular}
\end{table}

\subsection{Truth Recovery}
\label{sec:truth_recovery}
To assess the model's ability to reveal the original visual truth behind generated motion, we perform flow reversal experiments. This process aims to restore the underlying static content from motion dynamics, highlighting the forensic potential of our method. Since FoT defines a new proactive source-recovery setting for I2V, there is no existing deployable method that directly outputs the same recovered source image and template trajectory from generated evidence alone. We therefore use \textit{Forged} as a sanity reference, denoting the unprocessed I2V-generated frame or video, and report \textit{Ref.} as a privileged first-frame-anchor recovery. Specifically, \textit{Ref.} uses the first frame as an anchor and applies WAFT-DINOv3-a2~\cite{ofmodel-WAFT}, an optical-flow model requiring two input images, to estimate the correspondence needed for flow reversal and recovery. The \textit{Ref.} is included only to contextualize the effectiveness of FoT when such an anchor is available; it is not a competing deployable method in our proactive protection scenario, where an attacker would not preserve or provide the first-frame/source anchor together with the tampered video, and FoT operates from embedded template evidence alone.

\noindent\textbf{Evaluation Metrics.}
For evaluation, we compare pixel-level metrics (PSNR~\cite{metrics-PSNR}, SSIM~\cite{metrics-SSIM}), perceptual metrics (LPIPS~\cite{metrics-LPIPS}, pHash~\cite{metrics-pHash}), and semantic metrics (CLIP-Sim~\cite{CLIP}, DINO-Sim~\cite{metrics-DINOsim-oquab2023dinov2}) within the valid area, separately at the frame and video levels. Here, the valid area denotes pixels that can be reliably compared after source-to-frame alignment; pixels outside the image boundary, pixels left uncovered by scatter/warping, and pixels with invalid or ambiguous correspondence are excluded from pixel-wise metrics.

\begin{table}[!t]
\caption{Frame-level results of truth recovery. ``Forged'' denotes the unprocessed I2V-generated frame, and ``Ref.'' follows the definition in Sec.~\ref{sec:truth_recovery}. ``CLIP-S'' and ``DINO-S'' separately denote the ``CLIP-Sim'' and ``DINO-Sim''.}
\label{tab: frame-level recovery}
\resizebox{\linewidth}{!}{
\begin{tabular}{lcccccc}
\toprule
Settings & PSNR↑ & SSIM↑ & pHash↑ & LPIPS↓ & CLIP-S↑ & DINO-S↑ \\ \hline
\multicolumn{7}{c}{s0-10} \\ \hline
Forged & 20.3920 & 0.6380 & 0.9317 & 0.1256 & 0.9863 & 0.9760 \\
FoT & \textbf{23.0268} & \textbf{0.7585} & \textbf{0.9505} & \textbf{0.1186} & \textbf{0.9898} & \textbf{0.9823} \\
\textit{Ref.} & \textit{26.1460} & \textit{0.8346} & \textit{0.9727} & \textit{0.0984} & \textit{0.9942} & \textit{0.9921} \\ \hline
\multicolumn{7}{c}{s10-40} \\ \hline
Forged & 17.5963 & 0.6212 & 0.8486 & 0.2040 & 0.9658 & 0.9385 \\
FoT & \textbf{19.7398} & \textbf{0.7210} & \textbf{0.8845} & \textbf{0.1677} & \textbf{0.9719} & \textbf{0.9478} \\
\textit{Ref.} & \textit{24.7984} & \textit{0.8403} & \textit{0.9516} & \textit{0.1051} & \textit{0.9874} & \textit{0.9808} \\ \hline
\multicolumn{7}{c}{s40+} \\ \hline
Forged & 19.9284 & 0.7222 & 0.8037 & 0.2112 & 0.9449 & 0.9178 \\
FoT & \textbf{20.6162} & \textbf{0.7482} & \textbf{0.8186} & \textbf{0.1938} & \textbf{0.9474} & \textbf{0.9184} \\
\textit{Ref.} & \textit{26.4588} & \textit{0.8624} & \textit{0.9363} & \textit{0.0934} & \textit{0.9781} & \textit{0.9741} \\ \hline
\end{tabular}
}
\end{table}

\noindent\textbf{Frame-Level.}
As shown in Table~\ref{tab: frame-level recovery}, our method (FoT) improves over the Forged reference across all evaluated motion scales (s0-10, s10-40, and s40+). The improvement is not limited to pixel-level metrics; FoT also improves perceptual (LPIPS, pHash) and semantic (CLIP-Sim, DINO-Sim) similarity, moving the recovered content toward the privileged \textit{Ref.} result. Bold values compare FoT against Forged only; \textit{Ref.} is italicized as a reference row.

\begin{table}[!t]
\centering
\setlength{\abovecaptionskip}{5pt}
\caption{Video-level results of truth recovery. ``Forged'' denotes the unprocessed I2V-generated video evidence, and ``Ref.'' follows the definition in Sec.~\ref{sec:truth_recovery}.}
\label{tab: video-level recovery}
\resizebox{0.93\linewidth}{!}{
\begin{tabular}{lcccccc}
\toprule
Settings & PSNR↑ & SSIM↑ & pHash↑ & LPIPS↓ & CLIP-S↑ & DINO-S↑ \\ \midrule
\multicolumn{7}{c}{Drop 0\%} \\ \hline
Forged & 16.5468 & 0.5357 & 0.7662 & 0.4586 & 0.8676 & 0.6655 \\
FoT & \textbf{19.8573} & \textbf{0.6997} & \textbf{0.8611} & \textbf{0.2776} & \textbf{0.9074} & \textbf{0.8061} \\
\textit{Ref.} & \textit{19.3406} & \textit{0.7385} & \textit{0.8441} & \textit{0.2158} & \textit{0.9206} & \textit{0.8998} \\ \hline
\multicolumn{7}{c}{Drop 30\%} \\ \hline
Forged & 15.7337 & 0.5402 & 0.7691 & 0.3899 & 0.9159 & 0.7724 \\
FoT & \textbf{17.4325} & \textbf{0.6271} & \textbf{0.8093} & \textbf{0.3155} & \textbf{0.9288} & \textbf{0.8192} \\
\textit{Ref.} & \textit{18.9952} & \textit{0.7152} & \textit{0.8639} & \textit{0.2308} & \textit{0.9193} & \textit{0.8985} \\ \hline
\multicolumn{7}{c}{Drop 60\%} \\ \hline
Forged & 15.9234 & 0.5835 & 0.7864 & 0.3346 & 0.9372 & 0.8307 \\
FoT & \textbf{17.3381} & \textbf{0.6492} & \textbf{0.8169} & \textbf{0.2892} & \textbf{0.9412} & \textbf{0.8502} \\
\textit{Ref.} & \textit{19.5864} & \textit{0.7381} & \textit{0.8820} & \textit{0.2136} & \textit{0.9412} & \textit{0.9163} \\ \hline
\multicolumn{7}{c}{Drop 90\%} \\ \hline
Forged & 17.2483 & 0.6573 & 0.8096 & 0.2481 & 0.9497 & 0.9034 \\
FoT & \textbf{18.6647} & \textbf{0.7146} & \textbf{0.8350} & \textbf{0.2192} & \textbf{0.9508} & \textbf{0.9069} \\
\textit{Ref.} & \textit{22.1291} & \textit{0.8079} & \textit{0.9121} & \textit{0.1491} & \textit{0.9641} & \textit{0.9547} \\ \bottomrule
\end{tabular}
}
\end{table}

\begin{figure*}[t!]
    \centering
    \includegraphics[width=0.92\linewidth]{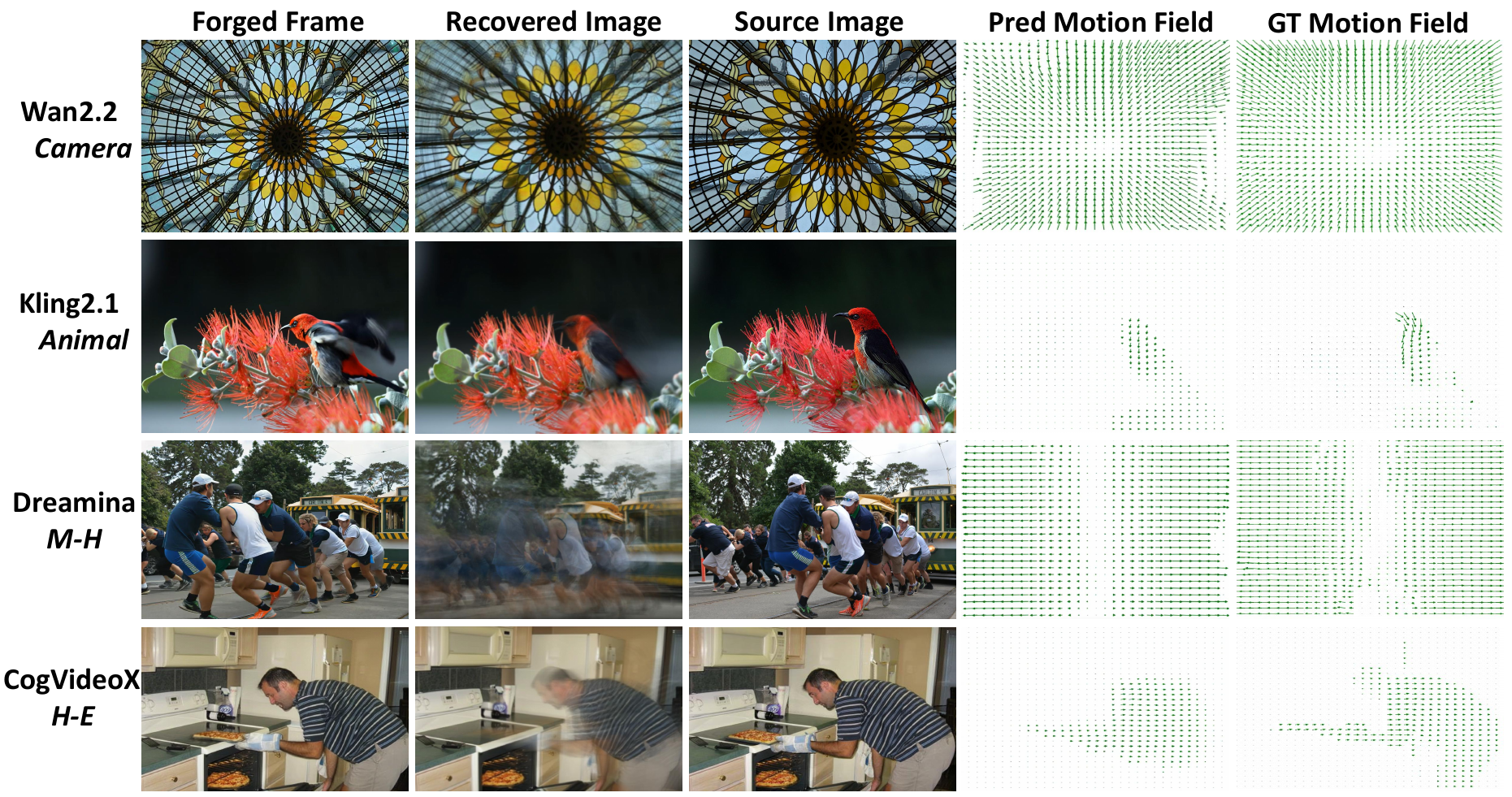}
    \caption{Qualitative results of video-level recovery. Our FoT module restores the original source image (Source Image) from the forged frame (Forged Frame) by predicting the underlying motion field (Pred Motion Field), which is compared with the pseudo-GT motion field estimated for evaluation. Wan2.2, Kling2.1, Dreamina, and CogVideoX denote I2V generator implementations rather than attack types.}
    \label{fig:Video-level Recovery}
\end{figure*}
\noindent\textbf{Video-Level.}
We further assess recovery robustness at the video level under initial-frame-dropping scenarios. Table~\ref{tab: video-level recovery} presents these results. Even when subjected to the extreme condition where the \textbf{first 90\% of frames are dropped}, FoT maintains a clear performance advantage over Forged, although a gap to the \textit{Ref.} row remains. This result indicates that FoT can still preserve useful forensic evidence under severe temporal data loss, while also showing that extreme frame removal remains a challenging case for source recovery.

\subsection{Qualitative Results}
Figure~\ref{fig:Video-level Recovery} shows the visual results of motion capture and flow reversal of FoT. Qualitatively, the FoT module shows a clear capability for \textbf{truth recovery} from frames forged by advanced video synthesis models.

\noindent\textbf{Visualization Protocol.}
For each qualitative case, we inspect the source image, the generated forged video, the frame-level recovered sequence, the video-level aggregated recovery, the predicted motion field, the confidence map, and the decoded template response.
This protocol is designed to evaluate not only whether the final recovered image resembles the source, but also whether the intermediate motion evidence remains temporally consistent across generated frames.
Table~\ref{tab:qualitative_protocol} summarizes the role of each diagnostic component.
\begin{table}[t]
\centering
\caption{Qualitative diagnostic components used to inspect FoT beyond scalar metrics.}
\label{tab:qualitative_protocol}
\footnotesize
\setlength{\tabcolsep}{0.04in}
\begin{tabular}{p{0.32\linewidth}p{0.60\linewidth}}
\toprule
\textbf{Component} & \textbf{Diagnostic Purpose} \\ \midrule
Source image & Provides the static visual truth that the recovery should reconstruct. \\
Forged video & Shows the semantic and temporal transformation introduced by the I2V generator. \\
Frame-level recovery & Tests whether each generated frame independently preserves enough forensic evidence for reversal. \\
Video-level recovery & Tests whether multiple recovered frames can be aggregated into a stable source estimate after frame dropping. \\
Motion arrows & Visualizes local displacement direction and exposes spatial discontinuities in predicted flow. \\
Confidence map & Indicates which regions are considered reliable by the motion prediction branch. \\
RGB flow & Encodes motion direction and magnitude in a dense form for comparison with pseudo ground truth. \\
Decoded template & Verifies whether the embedded proactive template can still be recovered after I2V generation. \\
\bottomrule
\end{tabular}
\end{table}

The frame-level recovery verifies per-frame reversibility, while the video-level recovery tests whether multiple noisy frame-wise estimates can be fused into a stable estimate when earlier frames are removed. FoT estimates motion from each available generated frame back to the source coordinate system independently, rather than chaining adjacent-frame flow; this design avoids temporal error accumulation when early frames are missing.
The motion-arrow and RGB-flow visualizations expose whether the predicted displacement direction and magnitude are spatially coherent, and the confidence map highlights where the model considers the recovered motion reliable.
Finally, comparing the decoded template with the originally embedded template provides a direct visual check of whether the forensic template survives the I2V generation process.
In practice, these visual components play complementary diagnostic roles.
The forged video reveals the semantic transformation imposed by the I2V generator, whereas the recovered frame sequence shows whether FoT can reverse this transformation without relying on a single lucky frame.
The aggregated recovery is particularly important for long videos because the first available frame may be absent or heavily distorted; therefore, stable recovery under progressive frame dropping indicates that the forensic template is distributed across the temporal trajectory rather than tied to a fixed frame index.
The motion-field views then separate motion failure from reconstruction failure: if the recovered image is degraded while the predicted flow remains coherent, the bottleneck is likely the reversal stage; if the flow itself becomes noisy or discontinuous, the limitation lies in template-guided motion capture.
This diagnostic protocol is used to interpret the quantitative trends in Tables~\ref{tab: frame-level recovery}, \ref{tab: video-level recovery}, and~\ref{tab: motion capture}.

The core observation lies in the content fidelity achieved by the restoration process. Across the examples, the ``Recovered Image'' is visibly closer to the ``Source Image'' than the ``Forged Frame''. This visual trend is consistent with the quantitative metrics in our frame-level and video-level recovery experiments (Tables~\ref{tab: frame-level recovery} and \ref{tab: video-level recovery}).

The success of the visual recovery is closely tied to the quality of flow prediction. The ``Pred Motion Field'' captures the main flow dynamics and remains close to the pseudo-GT motion field across complex scenes. This suggests that the FoT forensic template can separate the underlying scene motion from the forged content, which is the basis for reliable content restoration.

\subsection{Motion Capture}
\begin{figure}[!t]
\setlength{\abovecaptionskip}{3pt}
\captionsetup[subfloat]{font=footnotesize}
    \centering
    \subfloat[Across I2V models.\label{fig:sub1}]{
        \includegraphics[width=0.47\linewidth]{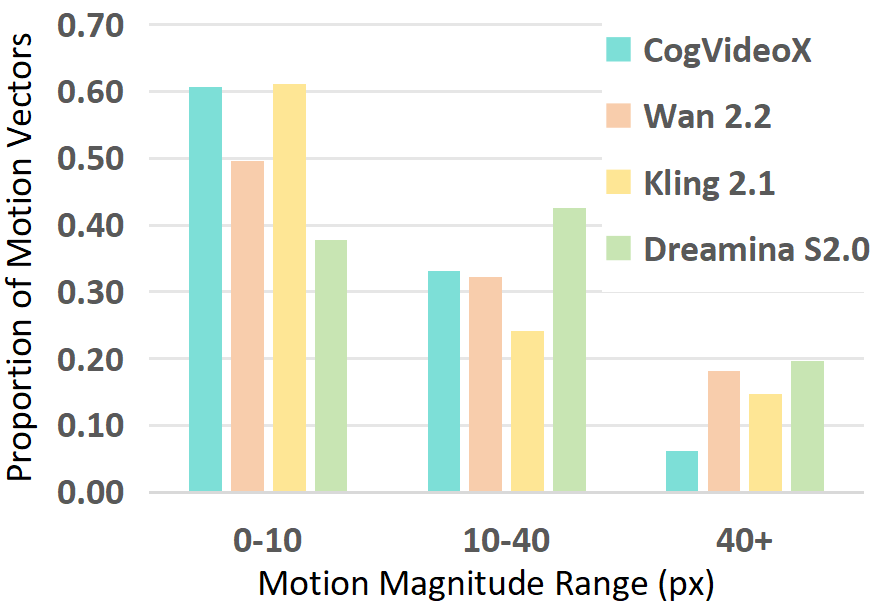}
    }
    \hfill
    \subfloat[Across scenarios.\label{fig:sub2}]{
        \includegraphics[width=0.47\linewidth]{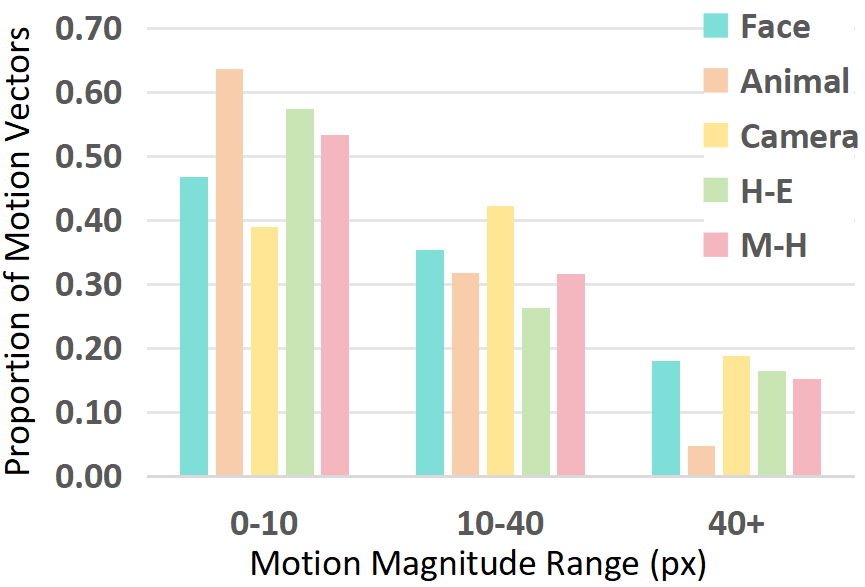}
    }
    \caption{Distribution of motion magnitude.}
    \label{fig:motion_distribution}
\end{figure}

\noindent\textbf{Evaluation Metrics.}
We adopt standard optical-flow metrics to quantify motion capture accuracy. 
For each generated video, dense forward ($\mathbf{F}_\text{fw}$) and backward ($\mathbf{F}_\text{bw}$) motion fields are estimated by the state-of-the-art WAFT-DINOv3-a2 model~\cite{ofmodel-WAFT}, using the $\mathbf{F}_\text{fw}$ as pseudo ground truth. This pseudo ground truth is used only for motion-capture evaluation; it is separate from the \textit{Ref.} recovery rows in Tables~\ref{tab: frame-level recovery} and~\ref{tab: video-level recovery}, although both use WAFT as a two-image optical-flow estimator. The pseudo ground truth may still contain errors in occluded or semantically re-synthesized regions. 
Following ARFlow~\cite{ofmodel-ARFlow}, a forward-backward consistency check determines valid (non-occluded) pixels by first warping the backward flow with $\mathbf{F}_\text{fw}$ and retaining pixels that satisfy
\begin{equation}
\mathbf{F}_\text{bw}^{\text{bwd}} = \mathcal{W}(\mathbf{F}_\text{bw}, \mathbf{F}_\text{fw}),
\end{equation}
\begin{equation}
\|\mathbf{F}_\text{fw} + \mathbf{F}_\text{bw}^{\text{bwd}}\|_2^2
\le 0.01(\|\mathbf{F}_\text{fw}\|_2^2 + \|\mathbf{F}_\text{bw}^{\text{bwd}}\|_2^2) + 0.5 .
\end{equation}

We first compute the pixel-wise \textbf{Endpoint Error (EPE)} between the pseudo ground-truth motion vector $\mathbf{v}_\text{gt}{=}(u_\text{gt},v_\text{gt})$ and the prediction $\mathbf{v}_\text{pred}{=}(u_\text{pred},v_\text{pred})$:
\begin{equation}
\text{EPE}=\sqrt{(u_\text{gt}-u_\text{pred})^2+(v_\text{gt}-v_\text{pred})^2}.
\end{equation}
The sample-wise average is reported as \textbf{AEE}:
\begin{equation}
\text{AEE}=\frac{\sum \text{EPE}\cdot M_\text{valid}}{\sum M_\text{valid}+\epsilon}.
\end{equation}
We also compute \textbf{Angle Error (AE)} and report its average as \textbf{AAE}:
\begin{equation}
\text{AAE}=
\frac{\sum \arccos \left(
\frac{\mathbf{v}_\text{pred}\cdot\mathbf{v}_\text{gt}}
{\max(\|\mathbf{v}_\text{pred}\|\|\mathbf{v}_\text{gt}\|,\epsilon)}
\right)\cdot M_\text{valid}}
{\sum M_\text{valid}+\epsilon}\cdot\frac{180}{\pi}.
\end{equation}
\textbf{Fl-all} follows KITTI~\cite{datasets-of-KITTI}, measuring the percentage of large-motion outliers:
\begin{equation}
\text{Fl-all}=
\frac{\sum \mathbf{1}\left((\text{EPE}>3)\wedge
(\text{EPE}/\|\mathbf{v}_\text{gt}\|>0.05)\right)\cdot M_\text{valid}}
{\sum M_\text{valid}+\epsilon}.
\end{equation}
\textbf{AUC} follows MemFlow~\cite{ofmodel-MemFlow}, integrating the inlier ratio over EPE thresholds from $0$ to $5$ px:
\begin{equation}
\text{AUC}=\int_0^5(1-\text{OutlierRate}(t))\,dt .
\end{equation}
Following Spring~\cite{datasets-of-Spring} and Sintel~\cite{datasets-of-Sintel}, we additionally report the \textbf{$n$-px outlier rate (R@n)} ($n{=}1,3,5$). 
For scale-aware analysis, motion magnitudes are grouped into \textit{0--10 px} (slow), \textit{10--40 px} (medium), and \textit{$>$40 px} (fast), following the common optical-flow practice of separating small, medium, and large displacements for diagnostic evaluation. All metrics are averaged over valid pixels within each range. 
For a video with $n$ frames, all $n$ frame pairs are included in the evaluation. The illustrated metrics are weighted averages calculated by the quantity of generated samples for each model or scenario.

\begin{table*}[!t]
\centering
\caption{
Quantitative results of our \textbf{FoT} model on motion capture across four I2V generators and five motion scenarios. 
Metrics include AEE, AAE, Fl-all, AUC, and per-scale ($s0$–$10$, $s10$–$40$, $s40+$) outlier rates at 1/3/5-pixel thresholds. 
Higher AUC and lower AEE / AAE / outlier rates indicate better motion reconstruction performance.
}
\label{tab: motion capture}
\resizebox{\textwidth}{!}{
\begin{tabular}{lccccccccccccc}
\toprule
\multicolumn{1}{c}{\multirow{2}{*}{\textbf{Scenarios}}} &
  \multirow{2}{*}{\textbf{AEE↓}} &
  \multirow{2}{*}{\textbf{AAE↓}} &
  \multirow{2}{*}{\textbf{Fl-all↓}} &
  \multicolumn{1}{c|}{\multirow{2}{*}{\textbf{AUC↑}}} &
  \multicolumn{3}{c|}{\textbf{s0-10}} &
  \multicolumn{3}{c|}{\textbf{s10-40}} &
  \multicolumn{3}{c}{\textbf{s40+}} \\ \cline{6-14} 
\multicolumn{1}{c}{} &
   &
   &
   &
  \multicolumn{1}{c|}{} &
  \textbf{R@1↓} &
  \textbf{R@3↓} &
  \multicolumn{1}{c|}{\textbf{R@5↓}} &
  \textbf{R@1↓} &
  \textbf{R@3↓} &
  \multicolumn{1}{c|}{\textbf{R@5↓}} &
  \textbf{R@1↓} &
  \textbf{R@3↓} &
  \textbf{R@5↓} \\ \hline
\multicolumn{14}{c}{\textbf{CogVideoX}} \\ \midrule
Face &
  4.7357 &
  23.5304 &
  0.1921 &
  \multicolumn{1}{c|}{0.7298} &
  0.4163 &
  0.1495 &
  \multicolumn{1}{c|}{0.0893} &
  0.5580 &
  0.3235 &
  \multicolumn{1}{c|}{0.2573} &
  0.2850 &
  0.2718 &
  0.2636 \\
Camera &
  6.5027 &
  26.8640 &
  0.2665 &
  \multicolumn{1}{c|}{0.7065} &
  0.4121 &
  0.1513 &
  \multicolumn{1}{c|}{0.1008} &
  0.5929 &
  0.3675 &
  \multicolumn{1}{c|}{0.2922} &
  0.2548 &
  0.2413 &
  0.2321 \\
Animal &
  5.5203 &
  22.3939 &
  0.1970 &
  \multicolumn{1}{c|}{0.7389} &
  0.3329 &
  0.1204 &
  \multicolumn{1}{c|}{0.0837} &
  0.6136 &
  0.4579 &
  \multicolumn{1}{c|}{0.4037} &
  0.3763 &
  0.3684 &
  0.3651 \\
H-E &
  8.7720 &
  27.4424 &
  0.2810 &
  \multicolumn{1}{c|}{0.7052} &
  0.3805 &
  0.1736 &
  \multicolumn{1}{c|}{0.1227} &
  0.6870 &
  0.5176 &
  \multicolumn{1}{c|}{0.4690} &
  0.5930 &
  0.5692 &
  0.5620 \\
M-H &
  8.0165 &
  24.7596 &
  0.3020 &
  \multicolumn{1}{c|}{0.7214} &
  0.4603 &
  0.2198 &
  \multicolumn{1}{c|}{0.1605} &
  0.7466 &
  0.5263 &
  \multicolumn{1}{c|}{0.4532} &
  0.6066 &
  0.5857 &
  0.5774 \\ \hline
\multicolumn{14}{c}{\textbf{Wan 2.2}} \\ \hline
Face &
  29.9811 &
  58.2631 &
  0.6283 &
  \multicolumn{1}{c|}{0.5884} &
  0.5229 &
  0.3485 &
  \multicolumn{1}{c|}{0.2773} &
  0.8056 &
  0.6959 &
  \multicolumn{1}{c|}{0.6489} &
  0.7392 &
  0.7204 &
  0.7130 \\
Camera &
  10.9887 &
  31.6956 &
  0.3522 &
  \multicolumn{1}{c|}{0.6810} &
  0.4240 &
  0.1969 &
  \multicolumn{1}{c|}{0.1426} &
  0.5952 &
  0.3953 &
  \multicolumn{1}{c|}{0.3366} &
  0.3686 &
  0.3375 &
  0.3219 \\
Animal &
  10.7740 &
  39.4060 &
  0.2638 &
  \multicolumn{1}{c|}{0.7831} &
  0.3544 &
  0.1597 &
  \multicolumn{1}{c|}{0.1162} &
  0.7322 &
  0.6165 &
  \multicolumn{1}{c|}{0.5743} &
  0.5187 &
  0.5178 &
  0.5173 \\
H-E &
  46.3807 &
  70.5741 &
  0.6628 &
  \multicolumn{1}{c|}{0.7020} &
  0.3227 &
  0.2325 &
  \multicolumn{1}{c|}{0.1957} &
  0.7017 &
  0.6524 &
  \multicolumn{1}{c|}{0.6298} &
  0.7887 &
  0.7871 &
  0.7862 \\
M-H &
  33.3723 &
  52.8603 &
  0.7609 &
  \multicolumn{1}{c|}{0.5417} &
  0.3506 &
  0.2376 &
  \multicolumn{1}{c|}{0.2012} &
  0.7719 &
  0.6811 &
  \multicolumn{1}{c|}{0.6490} &
  0.8085 &
  0.7994 &
  0.7951 \\ \hline
\multicolumn{14}{c}{\textbf{Kling 2.1}} \\ \hline
Face &
  19.1145 &
  49.4222 &
  0.5185 &
  \multicolumn{1}{c|}{0.6117} &
  0.4080 &
  0.2355 &
  \multicolumn{1}{c|}{0.1672} &
  0.8632 &
  0.7447 &
  \multicolumn{1}{c|}{0.6765} &
  0.6185 &
  0.6077 &
  0.6006 \\
Camera &
  22.5662 &
  43.4401 &
  0.5561 &
  \multicolumn{1}{c|}{0.6122} &
  0.5007 &
  0.2913 &
  \multicolumn{1}{c|}{0.2182} &
  0.6349 &
  0.4744 &
  \multicolumn{1}{c|}{0.4159} &
  0.5883 &
  0.5432 &
  0.5203 \\
Animal &
  6.7002 &
  71.2383 &
  0.2200 &
  \multicolumn{1}{c|}{0.7731} &
  0.2976 &
  0.1529 &
  \multicolumn{1}{c|}{0.1108} &
  0.8070 &
  0.7590 &
  \multicolumn{1}{c|}{0.7361} &
  0.4480 &
  0.4480 &
  0.4480 \\
H-E &
  28.5301 &
  79.4546 &
  0.4817 &
  \multicolumn{1}{c|}{0.7519} &
  0.3337 &
  0.2020 &
  \multicolumn{1}{c|}{0.1688} &
  0.7594 &
  0.7250 &
  \multicolumn{1}{c|}{0.7107} &
  0.7686 &
  0.7621 &
  0.7589 \\
M-H &
  25.5147 &
  70.3330 &
  0.4994 &
  \multicolumn{1}{c|}{0.7709} &
  0.3866 &
  0.2587 &
  \multicolumn{1}{c|}{0.2128} &
  0.8527 &
  0.8101 &
  \multicolumn{1}{c|}{0.7901} &
  0.7973 &
  0.7930 &
  0.7911 \\ \hline
\multicolumn{14}{c}{\textbf{Dreamina S2.0}} \\ \hline
Face &
  24.9238 &
  47.5357 &
  0.6965 &
  \multicolumn{1}{c|}{0.4338} &
  0.7198 &
  0.5136 &
  \multicolumn{1}{c|}{0.2949} &
  0.8017 &
  0.6362 &
  \multicolumn{1}{c|}{0.5583} &
  0.5358 &
  0.5240 &
  0.5151 \\
Camera &
  32.8325 &
  58.1933 &
  0.7538 &
  \multicolumn{1}{c|}{0.6751} &
  0.5640 &
  0.3986 &
  \multicolumn{1}{c|}{0.3267} &
  0.7728 &
  0.6660 &
  \multicolumn{1}{c|}{0.6226} &
  0.7016 &
  0.6916 &
  0.6837 \\
Animal &
  21.9378 &
  49.8771 &
  0.5990 &
  \multicolumn{1}{c|}{0.5859} &
  0.6211 &
  0.3900 &
  \multicolumn{1}{c|}{0.2916} &
  0.7728 &
  0.6166 &
  \multicolumn{1}{c|}{0.5562} &
  0.5909 &
  0.5880 &
  0.5842 \\
H-E &
  39.0917 &
  75.6922 &
  0.8194 &
  \multicolumn{1}{c|}{0.5724} &
  0.5631 &
  0.4404 &
  \multicolumn{1}{c|}{0.3349} &
  0.8256 &
  0.7733 &
  \multicolumn{1}{c|}{0.7489} &
  0.7813 &
  0.7795 &
  0.7790 \\
M-H &
  30.5991 &
  61.3336 &
  0.7864 &
  \multicolumn{1}{c|}{0.5849} &
  0.5879 &
  0.4454 &
  \multicolumn{1}{c|}{0.3470} &
  0.8930 &
  0.8290 &
  \multicolumn{1}{c|}{0.7906} &
  0.8118 &
  0.8086 &
  0.8065 \\ \bottomrule
\end{tabular}
}
\end{table*}


\begin{table*}[!t]
\centering
\caption{
Quantitative results of FoT on motion capture performance across four I2V models.
}
\label{tab: motion capture - across models}
\resizebox{\textwidth}{!}{
\setlength{\tabcolsep}{.1in}{
\begin{tabular}{lcccc|ccc|ccc|ccc}
\toprule
\multicolumn{1}{c}{\multirow{2}{*}{\textbf{I2V Models}}} &
  \multirow{2}{*}{\textbf{AEE↓}} &
  \multirow{2}{*}{\textbf{AAE↓}} &
  \multirow{2}{*}{\textbf{Fl-all↓}} &
  \multirow{2}{*}{\textbf{AUC↑}} &
  \multicolumn{3}{c|}{\textbf{s0-10}} &
  \multicolumn{3}{c|}{\textbf{s10-40}} &
  \multicolumn{3}{c}{\textbf{s40+}} \\ \cline{6-14} 
\multicolumn{1}{c}{} &
   &
   &
   &
   &
  \textbf{R@1↓} &
  \textbf{R@3↓} &
  \textbf{R@5↓} &
  \textbf{R@1↓} &
  \textbf{R@3↓} &
  \textbf{R@5↓} &
  \textbf{R@1↓} &
  \textbf{R@3↓} &
  \textbf{R@5↓} \\ \midrule
CogVideoX &
  6.8831 &
  25.1777 &
  0.2547 &
  0.7190 &
  0.4122 &
  0.1716 &
  0.1175 &
  0.6488 &
  0.4411 &
  0.3764 &
  0.4430 &
  0.4259 &
  0.4182 \\
Wan 2.2 &
  29.4018 &
  53.1208 &
  0.5908 &
  0.6341 &
  0.3998 &
  0.2494 &
  0.2007 &
  0.7310 &
  0.6235 &
  0.5843 &
  0.6848 &
  0.6719 &
  0.6660 \\
Kling 2.1 &
  22.3382 &
  62.6455 &
  0.4840 &
  0.7008 &
  0.3917 &
  0.2365 &
  0.1836 &
  0.7935 &
  0.7134 &
  0.6759 &
  0.6789 &
  0.6659 &
  0.6590 \\
Dreamina S2.0 &
  30.6163 &
  59.4429 &
  0.7484 &
  0.5607 &
  0.6139 &
  0.4478 &
  0.3230 &
  0.8243 &
  0.7220 &
  0.6740 &
  0.6981 &
  0.6919 &
  0.6874 \\ \bottomrule
\end{tabular}
}}
\end{table*}


\begin{table*}[!t]
\centering
\caption{
Quantitative results of FoT on motion capture performance across five threat scenarios.
}
\label{tab: motion capture - across scenarios}
\resizebox{\textwidth}{!}{
\setlength{\tabcolsep}{.1in}{
\begin{tabular}{lcccc|ccc|ccc|ccc}
\toprule
\multicolumn{1}{c}{\multirow{2}{*}{\textbf{Scenarios}}} &
  \multirow{2}{*}{\textbf{AEE↓}} &
  \multirow{2}{*}{\textbf{AAE↓}} &
  \multirow{2}{*}{\textbf{Fl-all↓}} &
  \multirow{2}{*}{\textbf{AUC↑}} &
  \multicolumn{3}{c|}{\textbf{s0-10}} &
  \multicolumn{3}{c|}{\textbf{s10-40}} &
  \multicolumn{3}{c}{\textbf{s40+}} \\ \cline{6-14} 
\multicolumn{1}{c}{} &
   &
   &
   &
   &
  \textbf{R@1↓} &
  \textbf{R@3↓} &
  \textbf{R@5↓} &
  \textbf{R@1↓} &
  \textbf{R@3↓} &
  \textbf{R@5↓} &
  \textbf{R@1↓} &
  \textbf{R@3↓} &
  \textbf{R@5↓} \\ \midrule
Face   & 20.6758 & 46.2935 & 0.5506 & 0.5621 & 0.5367 & 0.3384 & 0.2173 & 0.7890 & 0.6384 & 0.5700 & 0.5584 & 0.5457 & 0.5378 \\
Camera & 22.2362 & 44.6090 & 0.5553 & 0.6581 & 0.4994 & 0.2957 & 0.2290 & 0.6722 & 0.5158 & 0.4602 & 0.5489 & 0.5229 & 0.5083 \\
Animal & 12.5401 & 52.0093 & 0.3579 & 0.7030 & 0.4260 & 0.2336 & 0.1720 & 0.7562 & 0.6444 & 0.6009 & 0.4987 & 0.4964 & 0.4945 \\
H-E    & 32.0139 & 69.3399 & 0.5991 & 0.6741 & 0.4205 & 0.2871 & 0.2251 & 0.7642 & 0.7018 & 0.6778 & 0.7507 & 0.7441 & 0.7416 \\
M-H    & 25.9348 & 58.0442 & 0.6108 & 0.6645 & 0.4637 & 0.3165 & 0.2514 & 0.8401 & 0.7573 & 0.7214 & 0.7766 & 0.7696 & 0.7664 \\ \bottomrule
\end{tabular}
}}
\end{table*}

\begin{table}[!t]
\centering
\caption{Image quality and bit accuracy of watermarking methods under attacks when combined with the FoT module.
``FoT-pre'' indicates that FoT is used as a pre-embedder, while ``FoT-post'' indicates that FoT is used as a post-embedder.}
\label{tab:bit_wm}
\setlength{\tabcolsep}{1pt}
\renewcommand{\arraystretch}{1.08}
\begin{tabular}{@{}>{\raggedright\arraybackslash}p{0.13\linewidth}
>{\centering\arraybackslash}p{0.105\linewidth}
>{\centering\arraybackslash}p{0.105\linewidth}
>{\centering\arraybackslash}p{0.105\linewidth}
>{\centering\arraybackslash}p{0.105\linewidth}
>{\centering\arraybackslash}p{0.145\linewidth}
>{\centering\arraybackslash}p{0.145\linewidth}
>{\centering\arraybackslash}p{0.09\linewidth}@{}}
\toprule
\multirow{2}{*}{\textbf{Order}} &
\multicolumn{3}{c}{\textbf{Image Quality}} &
\multicolumn{4}{c}{\textbf{Bit Accuracy}} \\ \cmidrule(lr){2-4}\cmidrule(l){5-8}
& PSNR & SSIM & LPIPS & Clean & Degraded & Geometry & I2V \\ \midrule
\multicolumn{8}{c}{\textit{RoSteALS}~\cite{postprocessing_watermarking-Rosteals}} \\ \midrule
Native & 28.57 & 0.9261 & 0.0460 & \textbf{0.9841} & \underline{0.9669} & 0.5139 & 0.7324 \\
FoT-pre & 27.89 & 0.9062 & 0.0537 & \underline{0.9837} & \textbf{0.9718} & \underline{0.6881} & \underline{0.7866} \\
FoT-post & 27.89 & 0.9029 & 0.0554 & 0.9708 & 0.9571 & \textbf{0.8250} & \textbf{0.8291} \\ \midrule
\multicolumn{8}{c}{\textit{InvisMark}~\cite{postprocessing_watermarking-InvisMark}} \\ \midrule
Native & 49.68 & 0.9948 & 0.0005 & \textbf{0.8684} & \textbf{0.8507} & \underline{0.7303} & \underline{0.5399} \\
FoT-pre & 36.42 & 0.9672 & 0.0139 & \underline{0.8321} & \underline{0.8178} & \textbf{0.7902} & \textbf{0.5436} \\
FoT-post & 36.28 & 0.9663 & 0.0147 & 0.5413 & 0.5283 & 0.5272 & 0.5382 \\ \bottomrule
\end{tabular}
\end{table}

\begin{table*}[!t]
\centering
\caption{Quantitative results of tampering localization.}
\label{tab:localization}
\resizebox{\textwidth}{!}{
\begin{tabular}{lcccccccccccc}
\toprule
\multirow{2}{*}{\textbf{\begin{tabular}[c]{@{}l@{}}Settings\end{tabular}}} &
  \multicolumn{4}{c}{\textbf{ControlNet-Inpaint~\cite{DMs-ControlNet}}} &
  \multicolumn{4}{c}{\textbf{SDXL-1-Inpainting~\cite{ImageInpainting-SDXL}}} &
  \multicolumn{4}{c}{\textbf{SD-2-Inpainting~\cite{DMs-LDMs}}} \\ \cline{2-13} 
         & Dice   & IoU    & Pre    & \multicolumn{1}{c|}{Rec}    & Dice   & IoU    & Pre    & \multicolumn{1}{c|}{Rec}    & Dice   & IoU    & Pre    & Rec    \\ \midrule
Clean    & 0.9635 & 0.9337 & 0.9447 & \multicolumn{1}{c|}{0.9860} & 0.9679 & 0.9411 & 0.9513 & \multicolumn{1}{c|}{0.9877} & 0.9584 & 0.9250 & 0.9375 & 0.9845 \\
Degraded & 0.9215 & 0.8713 & 0.8909 & \multicolumn{1}{c|}{0.9731} & 0.9166 & 0.8676 & 0.8856 & \multicolumn{1}{c|}{0.9750} & 0.9091 & 0.8559 & 0.8784 & 0.9681 \\ \bottomrule
\end{tabular}
}
\end{table*}


\noindent\textbf{Detailed Quantitative Results.}
Table~\ref{tab: motion capture} reports the unaggregated motion-capture results for each I2V generator and each scenario before weighted averaging. 
This view complements Tables~\ref{tab: motion capture - across models} and~\ref{tab: motion capture - across scenarios}: instead of hiding per-scenario variations behind a single average, it exposes how each generator behaves under each forensic threat type.
The results show that CogVideoX remains consistently stable across all five scenarios, while Dreamina S2.0 and Wan~2.2 become substantially harder under interaction-heavy or large-motion cases. 
For example, Dreamina S2.0 reaches an AEE of 39.0917 in H-E and 30.5991 in M-H, and Wan~2.2 reaches an AEE of 29.9811 in Face and 33.3723 in M-H. 
Such raw results explain why the weighted summaries differ by both model and scenario: model-specific motion priors and scenario-specific motion structure jointly determine whether the embedded template can be accurately decoded into flow.

\noindent\textbf{Performance Analysis across I2V Models.}
Table~\ref{tab: motion capture - across models} and Fig.~\ref{fig:sub1} jointly reveal how FoT captures motion under different I2V models.
Table~\ref{tab: motion capture - across models} shows that FoT achieves its best motion-capture quality on CogVideoX (AEE~6.883, AUC~0.719), followed by Kling~2.1 and Wan~2.2. Dreamina~S2.0 yields the largest errors (AEE~30.616, AUC~0.561).

Combining these results with Fig.~\ref{fig:sub1}, we observe a strong correlation between performance differences and the distribution of true motion magnitudes. Models whose generated motions fall mostly within 0--40~px (e.g., CogVideoX and Kling~2.1) enable substantially more reliable motion capture. In contrast, Wan~2.2 produces noticeably larger motions, and Dreamina~S2.0 exhibits an even higher proportion of large displacements, where errors escalate.
This pattern echoes the long-standing challenge of large-motion estimation in optical flow. Our results indicate that this difficulty persists not only in image-based flow prediction but also in our template-guided flow prediction framework.

\noindent\textbf{Performance Analysis across Scenarios.}
Table~\ref{tab: motion capture - across scenarios} and Fig.~\ref{fig:sub2} together illustrate how FoT performs under different threat scenarios.
Table~\ref{tab: motion capture - across scenarios} shows that \textit{Animal} achieves the best results (AEE~12.5401, AUC~0.703), which aligns with its predominantly small-magnitude motion distribution (Fig.~\ref{fig:sub2}). 
However, despite having similarly low motion magnitudes, the \textit{H-E} and \textit{M-H} scenarios perform noticeably worse than both \textit{Face} and \textit{Camera}.
Interestingly, although \textit{Camera} exhibits the largest motion magnitudes overall, it still delivers the second-best performance (AEE~22.2362, Fl-all~0.5553, AUC~0.6581). 
This differs from the across-model observation, where larger motions degrade performance.
This is plausible as camera motion typically induces global, coherent displacement fields, which are easier to capture. In contrast, \textit{H-E} and \textit{M-H} involve complex 
interactions, producing nonrigid and spatially heterogeneous motion patterns. Thus, beyond motion magnitude, structural complexity also affects motion capture.

Taken together, the analyses across I2V models and across scenarios reveal two complementary factors governing FoT's performance: \textit{motion magnitude} and \textit{motion complexity}. 
Across models, performance is strongly tied to the distribution of motion magnitudes, with large motions consistently causing substantial errors. 
Across scenarios, however, motion magnitude alone cannot fully explain performance. Instead, structured and coherent motions (e.g., camera motion) remain tractable even when large, whereas complex, multi-agent, or highly deformable motions (e.g., \textit{H-E} or \textit{M-H}) pose greater challenges despite being smaller in magnitude. 
These findings indicate that FoT inherits the classical difficulty of large-motion estimation while additionally being sensitive to spatially complex motions, highlighting the need for future research on both large-range correspondence and fine-grained, nonrigid dynamics.

\noindent\textbf{Fine-Grained Observations from the Full Table.}
The detailed results in Table~\ref{tab: motion capture} further reveal several trends that are not visible from the weighted summaries alone.
First, CogVideoX provides the most stable testbed for FoT. Its AEE remains within a narrow range from 4.7357 on \textit{Face} to 8.7720 on \textit{H-E}, and its AUC stays around 0.70 across all scenarios. This indicates that CogVideoX tends to generate motion patterns that remain compatible with the learned forensic template, even when the semantic interaction becomes more complex.
The main degradation for CogVideoX appears in the high-motion outlier rates of \textit{H-E} and \textit{M-H}, suggesting that the model is not failing globally but is instead affected by localized regions where human interaction introduces nonrigid displacement.

Second, Wan~2.2 shows a much stronger dependence on scenario type. Although its \textit{Camera} case is relatively tractable (AEE~10.9887), its \textit{H-E} case reaches AEE~46.3807 and Fl-all~0.6628. This contrast supports the earlier interpretation that global camera motion is easier to model. Such content often contains both local body deformation and secondary object motion, making the pseudo ground-truth flow spatially heterogeneous. Under this condition, a single embedded template must account for multiple motion sources, which increases errors in both medium and fast motion ranges.

Third, Kling~2.1 exhibits a different profile. Its \textit{Animal} scenario has a low AEE of 6.7002 and a high AUC of 0.7731, but its AAE is high. This means the displacement magnitude can be recovered reasonably well while local direction may still fluctuate. Such behavior is consistent with small articulated animal motion, where the dominant displacement is limited but the direction varies over fine structures such as heads, wings, or limbs. For \textit{H-E} and \textit{M-H}, Kling~2.1 again becomes more difficult, with fast-motion outlier rates approaching 0.76--0.79, confirming that multi-region nonrigid dynamics remain a central challenge.

Finally, Dreamina S2.0 is the most challenging generator in the benchmark. It produces consistently high errors in interaction-heavy scenarios and also has high outlier rates in medium-motion regions. The \textit{H-E} and \textit{M-H} cases reach Fl-all values of 0.8194 and 0.7864, respectively, showing that the error is not limited to rare extreme pixels. Instead, a large portion of valid regions deviates substantially from the pseudo ground truth. This explains why Dreamina S2.0 has the weakest weighted performance even though some individual categories, such as \textit{Camera}, are not always the worst. Overall, these fine-grained results show that FoT's performance is governed by a three-way interaction among generator behavior, motion scale, and semantic structure.

\subsection{What Else Can FoT Do?}
\noindent
Beyond the proposed source-recovery and temporal-traceability setting, we also evaluate whether FoT can support two existing forensic tasks: \textbf{copyright authentication} and \textbf{tampering localization}. These tasks are parallel to our main task rather than direct downstream objectives. To assess FoT's zero-shot applicability, we freeze the FoT backbone and learnable template, keeping them task-agnostic.

\noindent\textbf{Enhancing Copyright Authentication.}
Geometric distortions are a long-standing weakness in watermarking. Prior work such as RoPaSS~\cite{postprocessing_watermarking-RoPaSS} mitigates this by embedding repetitive templates to recover rigid transformations. Inspired by this idea, we integrate our FoT as a \textit{plug-and-play} module to restore spatial alignment and facilitate watermark extraction. The results are illustrated in Table~\ref{tab:bit_wm}.

For RoSteALS, both integration orders improve robustness, but the post-embedding configuration is more effective. \textit{FoT-pre} increases bit accuracy under \textit{Geometric} (0.5139$\rightarrow$0.6881) and \textit{I2V} (0.7324$\rightarrow$0.7866) attacks, while \textit{FoT-post} further raises the scores to \textbf{0.8250} and \textbf{0.8291}, respectively. This indicates that FoT can act as a geometric recovery layer: after I2V or geometric distortion, the FoT branch helps restore spatial alignment before the watermark is decoded.

The behavior of InvisMark is different. \textit{FoT-pre} improves geometric robustness (0.7303$\rightarrow$\textbf{0.7902}) but barely changes I2V robustness (0.5399$\rightarrow$0.5436), and \textit{FoT-post} causes a severe drop even on clean inputs. This contrast suggests that the gain depends on both watermark-space compatibility and the native compression robustness of the partner watermark. RoSteALS already retains partial robustness to I2V compression, so alignment recovery by FoT can translate into better bit recovery. InvisMark, in contrast, has very high image fidelity but much weaker native I2V robustness, reflecting the common fidelity--robustness trade-off in watermarking; its compact embedding space appears more sensitive to distortions outside the training distribution and can be disrupted by an additional FoT embedding.

These results position FoT as a geometric-robustness enhancer rather than a universally plug-compatible watermark wrapper. Future watermark-specific integration is likely to further improve performance: FoT can be inserted as a distortion layer during watermark training, used in a serial embedding pipeline, or jointly optimized with the bit watermark and forensic template in a parallel embedding design. Such co-training would encourage the watermark payload and FoT template to occupy more compatible embedding spaces while benefiting from FoT's alignment recovery for geometric robustness.

\noindent\textbf{Intrinsic Tampering Localization Capability.}
While predicting motion is challenging, detecting content modifications in an image is relatively straightforward. Given that our forensic template effectively captures motion dynamics, we hypothesize that it can also aid in tampering localization. 

\begin{figure}[!t]
\captionsetup[subfloat]{font=footnotesize}
    \centering
    \subfloat[Protected]{\includegraphics[width=0.31\linewidth]{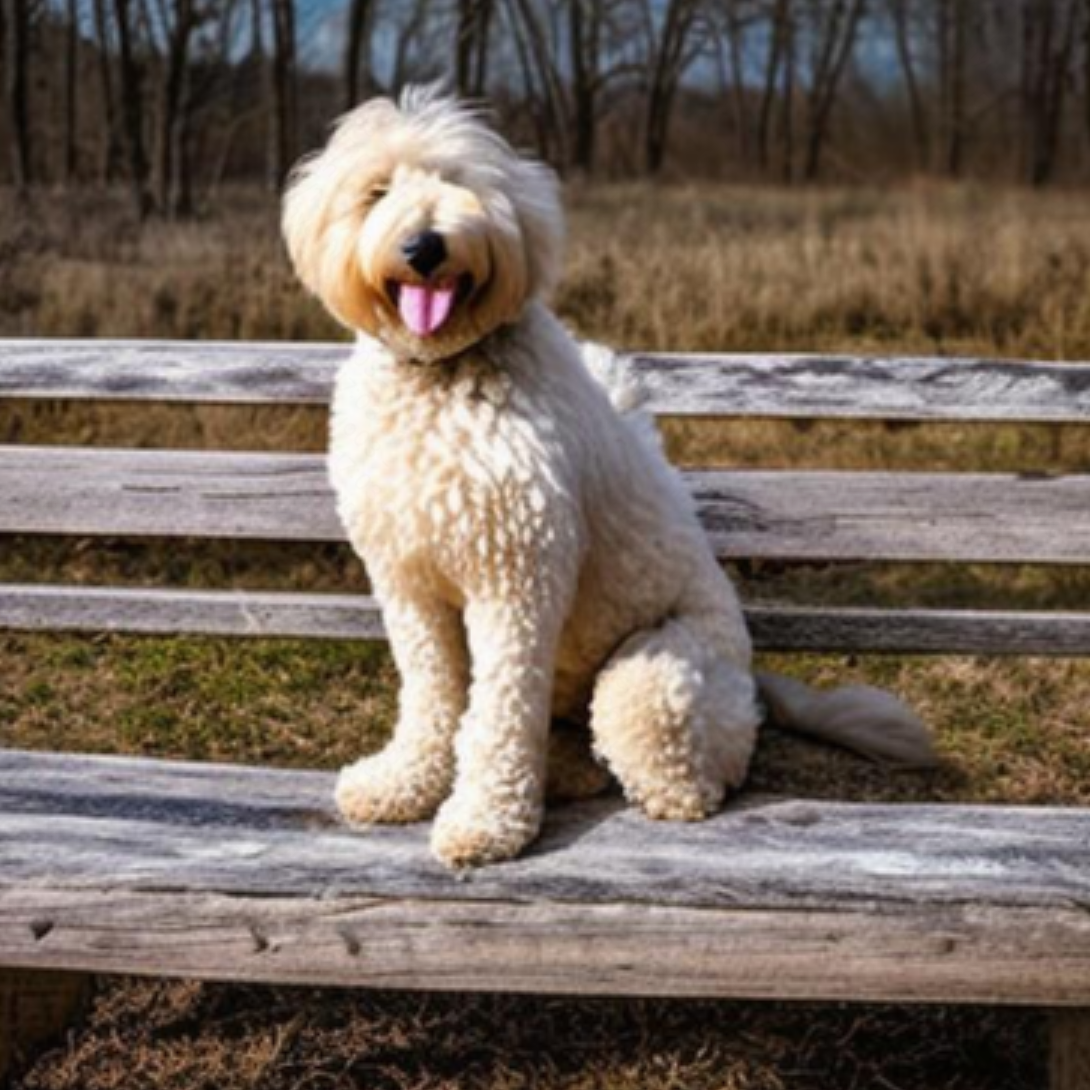}}
    \hfill
    \subfloat[Tampered]{\includegraphics[width=0.31\linewidth]{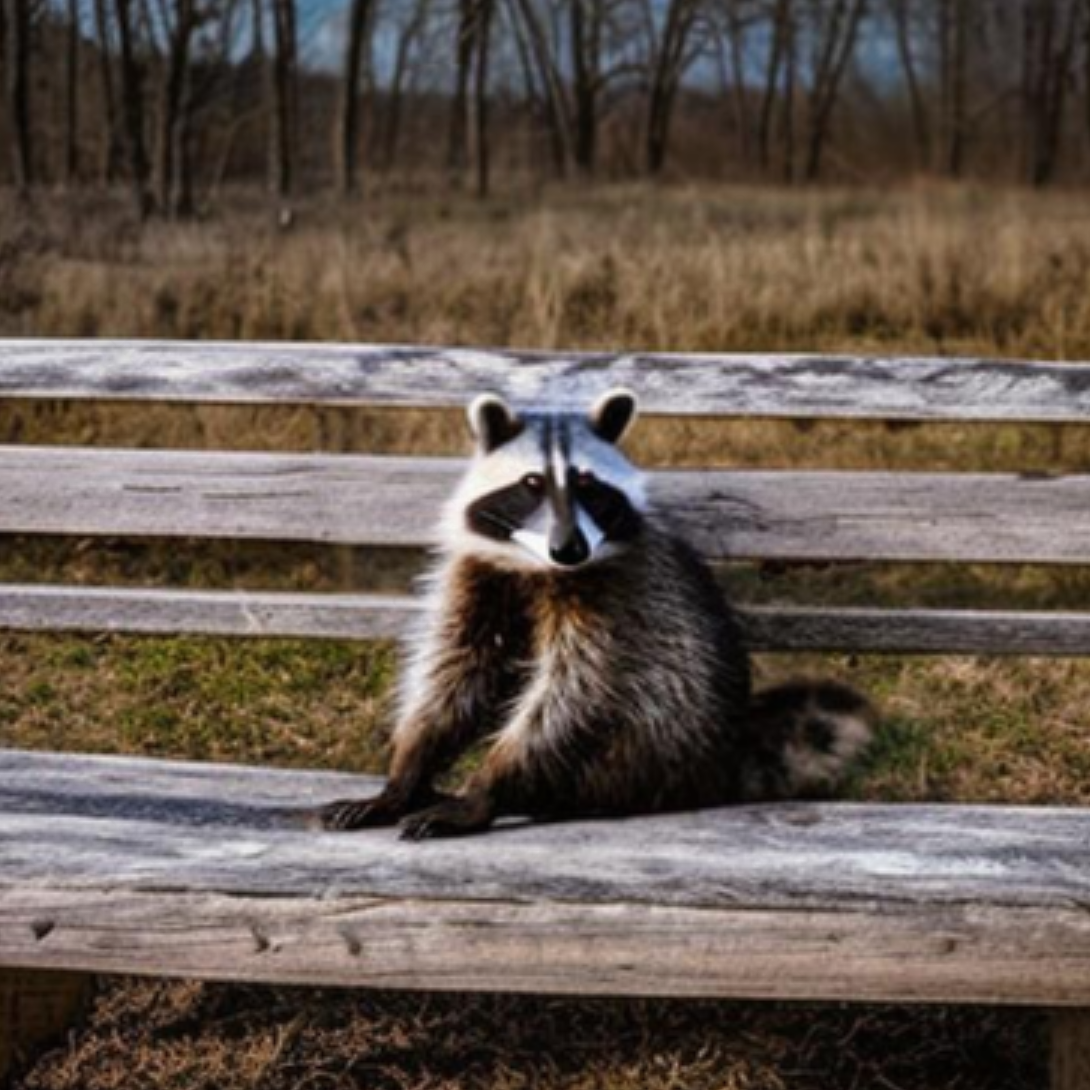}}
    \hfill
    \subfloat[Localization]{\includegraphics[width=0.31\linewidth]{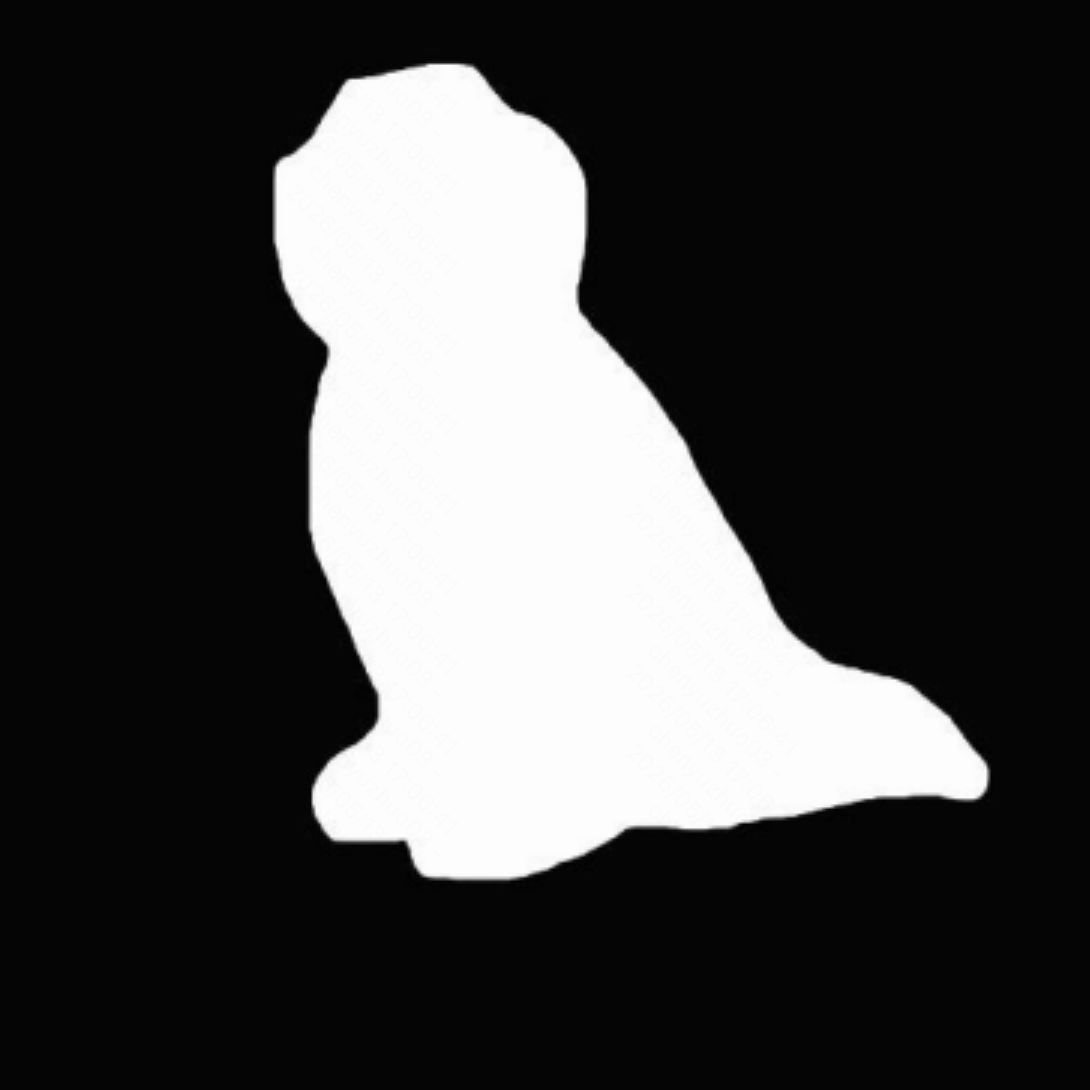}}
    \caption{Illustration of template-based tampering localization. A protected image is locally edited, and the localization branch localizes the edited region from template inconsistency.}
    \label{fig:tampering_localization_example}
\end{figure}

To test this hypothesis, we freeze all backbone parameters and attach a lightweight tampering localization head. Given an edited FoT-embedded image, the frozen decoder and backbone extract template-consistency features; regions whose decoded template response deviates from the expected embedded evidence are then mapped to a tampering mask by the localization head. The Content Predictor was trained exclusively for $108\text{K}$ steps, with \textbf{all other backbone parameters frozen}. The optimization objective comprised a combination of the \textit{Soft Dice loss} and the \textit{Binary Cross-Entropy (BCE) loss}, initialized with a learning rate of $1 \times 10^{-4}$ for optimization. We evaluate this capability on three local editing models.

Figure~\ref{fig:tampering_localization_example} gives a qualitative example of the localization workflow, and Table~\ref{tab:localization} reports the quantitative results on three distinct advanced inpainting models.
The results on \textbf{Clean} inputs are high, with Dice scores above $0.95$ across all models, validating the efficacy of the motion-based features for segmentation.
Even when confronted with \textbf{Degraded} inputs, our framework maintains high performance; for instance, the Dice score remains above $0.91$ on the challenging SDXL-1-Inpainting model.
We observe that the recall (Rec) metrics remain high in the Degraded setting, suggesting that the tampered regions are rarely missed. The slight drop in precision (Pre) indicates that the primary failure mode under degradation is the introduction of minimal false positives.
These findings support our hypothesis that features learned for motion dynamics can also generalize to the forensic task of content modification localization.

\section{Discussion}
\label{Discussion}

\noindent\textbf{When FoT Works Best.}
The full set of experiments suggests that FoT is most effective when the embedded evidence survives as a spatially coherent trace through the I2V pipeline.
This condition appears in two broad cases.
First, when the generated motion is small or moderate, the template remains close enough to its source position that the flow predictor can recover reliable correspondences, as reflected by the strong performance on CogVideoX and the \textit{Animal} scenario.
Second, even when the motion magnitude is large, FoT remains effective if the displacement field is globally structured, as in camera motion.
This explains why \textit{Camera} can outperform interaction-heavy categories despite having larger overall motion.
For deployment, these observations imply that forensic reliability should not be judged only by the amount of visible motion in a video.
A large but coherent camera movement can be easier to analyze than a smaller motion with deformable objects, occlusions, or human-object interactions.

\noindent\textbf{Failure Modes.}
The experiments also expose the main failure modes of proactive temporal forensics.
The most difficult cases are not simply those with high pixel displacement, but those where the I2V generator semantically rewrites the scene while introducing spatially heterogeneous motion.
In \textit{H-E} and \textit{M-H} videos, local body deformation, object interaction, and multiple moving subjects can break the assumption that a single embedded template can be smoothly transported through time. These cases also reveal the gap between our VAE-plus-flow simulation and real I2V behavior: semantic re-synthesis, occlusion, newly generated content, and highly specific motions such as faces or hands are only partially covered by the motion bank.
This limitation is consistent with the fine-grained results for Wan~2.2 and Dreamina~S2.0, where the degradation is visible not only in average endpoint error but also in the medium- and fast-motion outlier rates.
FoT uses uncertainty modeling to reduce the influence of such unreliable regions during fusion, but the current model still cannot fully recover content that has been semantically replaced. Therefore, future improvements should prioritize correspondence reasoning in nonrigid regions, rather than only increasing template strength.
Promising directions include multi-scale motion heads, region-aware confidence aggregation, and semantic priors that distinguish object deformation from camera-induced displacement. Targeted motion collection for face, hand, or interaction-heavy scenes could further improve these difficult categories, while the present work focuses on the base model design and a general-purpose motion bank.

\noindent\textbf{Implications for Existing Forensic Tasks.}
The auxiliary forensic experiments further clarify what FoT contributes beyond direct source recovery.
In watermarking, the effect of FoT is governed by both embedding-space compatibility and the partner watermark's native robustness to compression-like transformations.
RoSteALS benefits from both integration orders, with stronger gains in the FoT-post configuration, because its watermark can still survive I2V-like compression and then benefit from FoT's spatial realignment.
InvisMark is more fragile: its very high image fidelity indicates a tighter embedding space and a stronger fidelity--robustness trade-off, so additional FoT embedding or unseen I2V distortions can disrupt payload recovery.
This suggests that FoT is best understood as a geometric-robustness enhancer for watermarking, not as a drop-in replacement for watermark-specific robustness training.
Future systems could fine-tune watermarking methods with FoT as a distortion layer, serialize watermark and template embedding, or jointly embed the payload and FoT template so that the two signals are optimized to coexist.
In tampering localization, the same forensic template transfers well because localization only requires identifying inconsistent regions, not reconstructing every displaced pixel.
Together, these two auxiliary studies indicate that FoT can be viewed as a temporal forensic substrate: it can support recovery, authentication, and localization, but each task must respect the interaction between FoT's forensic template and the task-specific signal. In deployment, FoT is a proactive protection method: the source image must be embedded before publication or before entering an I2V workflow. Unprotected images are outside this proactive setting because they do not carry the template evidence needed by the recovery pipeline.

\noindent\textbf{Future Experimental Directions.}
The current evaluation applies one post-processing operation at a time to keep robustness loss interpretable, while the video-dropping experiment tests recovery under early-frame removal. Real platforms may combine recompression, resizing, cropping, and frame selection, so compounded degradations remain a natural next benchmark. Our current training normalizes inputs to $512\times512$, and arbitrary-resolution watermarking systems such as TrustMark~\cite{postprocessing_watermarking-TrustMark} suggest a scale-adaptive deployment path: normalize or partition images into embedding-compatible regions, then aggregate recovered payloads or template responses across scales. Model-adaptive stress tests, second-stage video-to-video editing, and multi-frame trajectory constraints are also useful future directions for clarifying when proactive temporal evidence remains reliable.

\section{Conclusion}
\label{conclusion}
In this paper, we introduced Flow of Truth (FoT), a proactive forensic framework designed for the temporal traceability challenge of image-to-video (I2V) generation. Instead of treating I2V generation only as frame synthesis, we model it from a pixel-motion perspective. By learning a forensic template that evolves with the underlying pixel flow, FoT can trace temporal manipulations back to their source evidence.
Our experiments show that FoT recovers the original source image from generated video frames and performs consistently across both commercial and open-source I2V models. The framework also remains effective when a large portion of video frames is discarded by an attacker. Beyond recovery, we show that FoT can enhance downstream forensic tasks, such as copyright watermarking and tampering localization.
At the same time, tracking pixels through large, non-rigid displacements and semantic reinterpretations remains challenging. Future work could explore high-level generative priors and attention-based reasoning to build a more complete model of temporal evolution. We hope this work provides a useful step toward proactive temporal forensic tools for the era of generative video.

\bibliographystyle{IEEEtran}
\bibliography{references}

@String(CVPR= {IEEE Conf. Comput. Vis. Pattern Recog.})

@String(ICCV= {Int. Conf. Comput. Vis.})

@String(ECCV= {Eur. Conf. Comput. Vis.})

@String(AAAI = {AAAI})

@String(CVPR  = {CVPR})

@String(ICCV  = {ICCV})

@String(ECCV  = {ECCV})

@article{videogen-AnimateDiff,
  title={Animatediff: Animate your personalized text-to-image diffusion models without specific tuning},
  author={Guo, Yuwei and Yang, Ceyuan and Rao, Anyi and Liang, Zhengyang and Wang, Yaohui and Qiao, Yu and Agrawala, Maneesh and Lin, Dahua and Dai, Bo},
  journal={arXiv preprint arXiv:2307.04725},
  year={2023}
}

@article{videogen-MovieGen,
  title={Movie gen: A cast of media foundation models},
  author={Polyak, Adam and Zohar, Amit and Brown, Andrew and Tjandra, Andros and Sinha, Animesh and Lee, Ann and Vyas, Apoorv and Shi, Bowen and Ma, Chih-Yao and Chuang, Ching-Yao and others},
  journal={arXiv preprint arXiv:2410.13720},
  year={2024}
}

@article{videogen-wan,
  title={Wan: Open and advanced large-scale video generative models},
  author={Wan, Team and Wang, Ang and Ai, Baole and Wen, Bin and Mao, Chaojie and Xie, Chen-Wei and Chen, Di and Yu, Feiwu and Zhao, Haiming and Yang, Jianxiao and others},
  journal={arXiv preprint arXiv:2503.20314},
  year={2025}
}

@article{videogen-Hunyuan,
  title={Hunyuanvideo: A systematic framework for large video generative models},
  author={Kong, Weijie and Tian, Qi and Zhang, Zijian and Min, Rox and Dai, Zuozhuo and Zhou, Jin and Xiong, Jiangfeng and Li, Xin and Wu, Bo and Zhang, Jianwei and others},
  journal={arXiv preprint arXiv:2412.03603},
  year={2024}
}

@article{videogen-CogVideoX,
  title={Cogvideox: Text-to-video diffusion models with an expert transformer},
  author={Yang, Zhuoyi and Teng, Jiayan and Zheng, Wendi and Ding, Ming and Huang, Shiyu and Xu, Jiazheng and Yang, Yuanming and Hong, Wenyi and Zhang, Xiaohan and Feng, Guanyu and others},
  journal={arXiv preprint arXiv:2408.06072},
  year={2024}
}

@misc{videogen-sora,
  author = {OpenAI},
  title = {Sora},
  howpublished = {\url{https://openai.com/sora}},
  year = {2024},
  note = {Accessed: 2025-11-08}
}

@misc{videogen-veo,
  author = {Google DeepMind},
  title = {Veo},
  howpublished = {\url{https://deepmind.google/veo/}},
  year = {2024},
  note = {Accessed: 2025-11-08}
}

@misc{videogen-kling,
  author = {Kuaishou Technology},
  title = {Kling},
  howpublished = {\url{https://klingai.com/}},
  year = {2024},
  note = {Accessed: 2025-11-08}
}

@misc{videogen-dreamina,
  author = {ByteDance},
  title = {Dreamina},
  howpublished = {\url{https://dreamina.capcut.com/}},
  year = {2024},
  note = {Accessed: 2025-11-08}
}

@inproceedings{ofmodel-SEA-RAFT,
  title={Sea-raft: Simple, efficient, accurate raft for optical flow},
  author={Wang, Yihan and Lipson, Lahav and Deng, Jia},
  booktitle={European Conference on Computer Vision},
  pages={36--54},
  year={2024},
  organization={Springer}
}

@article{ofmodel-WAFT,
  title={WAFT: Warping-Alone Field Transforms for Optical Flow},
  author={Wang, Yihan and Deng, Jia},
  journal={arXiv preprint arXiv:2506.21526
        
        
        
        
        
        
        
        },
  year={2025}
}

@inproceedings{ofmodel-MemFlow,
  title={MemFlow: Optical Flow Estimation and Prediction with Memory},
  author={Dong, Qiaole and Fu, Yanwei},
  booktitle={Proceedings of the IEEE/CVF Conference on Computer Vision and Pattern Recognition},
  year={2024}
}

@inproceedings{ofmodel-ARFlow,
   title = {Learning by Analogy: Reliable Supervision from Transformations for Unsupervised Optical Flow Estimation},
   author = {Liu, Liang and Zhang, Jiangning and He, Ruifei and Liu, Yong and Wang, Yabiao and Tai, Ying and Luo, Donghao and Wang, Chengjie and Li, Jilin and Huang, Feiyue},
   booktitle = {IEEE Conference on Computer Vision and Pattern Recognition(CVPR)},
   year = {2020}
}

@inproceedings{ofmodel-GMFlow,
  title={GMFlow: Learning Optical Flow via Global Matching},
  author={Xu, Haofei and Zhang, Jing and Cai, Jianfei and Rezatofighi, Hamid and Tao, Dacheng},
  booktitle={Proceedings of the IEEE/CVF Conference on Computer Vision and Pattern Recognition},
  pages={8121-8130},
  year={2022}
}

@inproceedings{ofmodel-FlowFormer++,
  title={Flowformer++: Masked cost volume autoencoding for pretraining optical flow estimation},
  author={Shi, Xiaoyu and Huang, Zhaoyang and Li, Dasong and Zhang, Manyuan and Cheung, Ka Chun and See, Simon and Qin, Hongwei and Dai, Jifeng and Li, Hongsheng},
  booktitle={Proceedings of the IEEE/CVF Conference on Computer Vision and Pattern Recognition},
  pages={1599--1610},
  year={2023}
}

@inproceedings{ofmodel-RAFT,
  title={Raft: Recurrent all-pairs field transforms for optical flow},
  author={Teed, Zachary and Deng, Jia},
  booktitle={European conference on computer vision},
  pages={402--419},
  year={2020},
  organization={Springer}
}

@inproceedings{DMs-LDMs,
  title={High-resolution image synthesis with latent diffusion models},
  author={Rombach, Robin and Blattmann, Andreas and Lorenz, Dominik and Esser, Patrick and Ommer, Bj{\"o}rn},
  booktitle={Proceedings of the IEEE/CVF conference on computer vision and pattern recognition},
  pages={10684--10695},
  url={https://arxiv.org/abs/2112.10752},
  year={2022}
}

@article{ImageInpainting-SDXL,
  title={Sdxl: Improving latent diffusion models for high-resolution image synthesis},
  author={Podell, Dustin and English, Zion and Lacey, Kyle and Blattmann, Andreas and Dockhorn, Tim and M{\"u}ller, Jonas and Penna, Joe and Rombach, Robin},
  journal={arXiv preprint arXiv:2307.01952},
  year={2023}
}

@inproceedings{DMs-ControlNet,
  title={Adding conditional control to text-to-image diffusion models},
  author={Zhang, Lvmin and Rao, Anyi and Agrawala, Maneesh},
  booktitle={Proceedings of the IEEE/CVF International Conference on Computer Vision},
  pages={3836--3847},
  year={2023}
}

@article{metrics-DINOsim-oquab2023dinov2,
  title={Dinov2: Learning robust visual features without supervision},
  author={Oquab, Maxime and Darcet, Timoth{\'e}e and Moutakanni, Th{\'e}o and Vo, Huy and Szafraniec, Marc and Khalidov, Vasil and Fernandez, Pierre and Haziza, Daniel and Massa, Francisco and El-Nouby, Alaaeldin and others},
  journal={arXiv preprint arXiv:2304.07193
        
        
        
        
        
        },
  year={2023}
}

@article{postprocessing_watermarking-HiDDeN,
  title={HiDDeN: hiding data with deep networks},
  author={Zhu, J},
  journal={arXiv preprint arXiv:1807.09937
        
        },
  year={2018}
}

@inproceedings{postprocessing_watermarking-Stegastamp,
  title={Stegastamp: Invisible hyperlinks in physical photographs},
  author={Tancik, Matthew and Mildenhall, Ben and Ng, Ren},
  booktitle={Proceedings of the IEEE/CVF conference on computer vision and pattern recognition},
  pages={2117--2126},
  year={2020}
}

@inproceedings{postprocessing_watermarking-TrustMark,
  title={TrustMark: Robust Watermarking and Watermark Removal for Arbitrary Resolution Images},
  author={Bui, Tu and Agarwal, Shruti and Collomosse, John},
  booktitle={Proceedings of the IEEE/CVF International Conference on Computer Vision},
  year={2025}
}

@inproceedings{postprocessing_watermarking-REVMark,
  title={A Novel Deep Video Watermarking Framework with Enhanced Robustness to H. 264/AVC Compression},
  author={Zhang, Yulin and Ni, Jiangqun and Su, Wenkang and Liao, Xin},
  booktitle={Proceedings of the 31st ACM International Conference on Multimedia},
  pages={8095--8104},
  year={2023}
}

@inproceedings{postprocessing_watermarking-RoPaSS,
  title={RoPaSS: Robust Watermarking for Partial Screen-Shooting Scenarios},
  author={Ma, Zehua and Fang, Han and Yang, Xi and Chen, Kejiang and Zhang, Weiming},
  booktitle={Proceedings of the AAAI Conference on Artificial Intelligence},
  volume={39},
  number={18},
  pages={19332--19339},
  year={2025}
}

@inproceedings{postprocessing_watermarking-Rosteals,
  title={Rosteals: Robust steganography using autoencoder latent space},
  author={Bui, Tu and Agarwal, Shruti and Yu, Ning and Collomosse, John},
  booktitle={Proceedings of the IEEE/CVF conference on computer vision and pattern recognition},
  pages={933--942},
  year={2023}
}

@inproceedings{postprocessing_watermarking-InvisMark,
  title={InvisMark: Invisible and Robust Watermarking for AI-generated Image Provenance},
  author={Xu, Rui and Hu, Mengya and Lei, Deren and Li, Yaxi and Lowe, David and Gorevski, Alex and Wang, Mingyu and Ching, Emily and Deng, Alex},
  booktitle={2025 IEEE/CVF Winter Conference on Applications of Computer Vision (WACV)},
  pages={909--918},
  year={2025},
  organization={IEEE}
}

@inproceedings{wm_and_tl-Editguard,
  title={Editguard: Versatile image watermarking for tamper localization and copyright protection},
  author={Zhang, Xuanyu and Li, Runyi and Yu, Jiwen and Xu, Youmin and Li, Weiqi and Zhang, Jian},
  booktitle={Proceedings of the IEEE/CVF Conference on Computer Vision and Pattern Recognition},
  pages={11964--11974},
  year={2024}
}

@article{wm_and_tl-OmniGuard,
  title={OmniGuard: Hybrid Manipulation Localization via Augmented Versatile Deep Image Watermarking},
  author={Zhang, Xuanyu and Tang, Zecheng and Xu, Zhipei and Li, Runyi and Xu, Youmin and Chen, Bin and Gao, Feng and Zhang, Jian},
  journal={arXiv preprint arXiv:2412.01615
        
        },
  year={2024}
}

@inproceedings{CLIP,
  title={Learning transferable visual models from natural language supervision},
  author={Radford, Alec and Kim, Jong Wook and Hallacy, Chris and Ramesh, Aditya and Goh, Gabriel and Agarwal, Sandhini and Sastry, Girish and Askell, Amanda and Mishkin, Pamela and Clark, Jack and others},
  booktitle={International conference on machine learning},
  pages={8748--8763},
  year={2021},
  organization={PMLR}
}

@inproceedings{metrics-PSNR,
  title={Stego image quality and the reliability of PSNR},
  author={Almohammad, Adel and Ghinea, Gheorghita},
  booktitle={2010 2nd International Conference on Image Processing Theory, Tools and Applications},
  pages={215--220},
  year={2010},
  organization={IEEE}
}

@article{metrics-SSIM,
  title={Image quality assessment: from error visibility to structural similarity},
  author={Wang, Zhou and Bovik, Alan C and Sheikh, Hamid R and Simoncelli, Eero P},
  journal={IEEE transactions on image processing},
  volume={13},
  number={4},
  pages={600--612},
  year={2004},
  publisher={IEEE}
}

@inproceedings{metrics-LPIPS,
  title={The unreasonable effectiveness of deep features as a perceptual metric},
  author={Zhang, Richard and Isola, Phillip and Efros, Alexei A and Shechtman, Eli and Wang, Oliver},
  booktitle={Proceedings of the IEEE conference on computer vision and pattern recognition},
  pages={586--595},
  year={2018}
}

@article{metrics-pHash,
  title={Analysis of perceptual hashing algorithms in image manipulation detection},
  author={Samanta, Priyanka and Jain, Shweta},
  journal={Procedia Computer Science},
  volume={185},
  pages={203--212},
  year={2021},
  publisher={Elsevier}
}

@inproceedings{datasets-DIV2K,
  title={Ntire 2017 challenge on single image super-resolution: Dataset and study},
  author={Agustsson, Eirikur and Timofte, Radu},
  booktitle={Proceedings of the IEEE conference on computer vision and pattern recognition workshops},
  pages={126--135},
  year={2017}
}

@inproceedings{datasets-FFHQ,
  title={A style-based generator architecture for generative adversarial networks},
  author={Karras, Tero and Laine, Samuli and Aila, Timo},
  booktitle={Proceedings of the IEEE/CVF conference on computer vision and pattern recognition},
  pages={4401--4410},
  year={2019}
}

@inproceedings{datasets-MSCOCO,
  title={Microsoft coco: Common objects in context},
  author={Lin, Tsung-Yi and Maire, Michael and Belongie, Serge and Hays, James and Perona, Pietro and Ramanan, Deva and Doll{\'a}r, Piotr and Zitnick, C Lawrence},
  booktitle={European conference on computer vision},
  pages={740--755},
  year={2014},
  organization={Springer}
}

@inproceedings{datasets-of-KITTI,
  author = {Moritz Menze and Andreas Geiger},
  title = {Object Scene Flow for Autonomous Vehicles},
  booktitle = {Conference on Computer Vision and Pattern Recognition (CVPR)},
  year = {2015}
}

@inproceedings{datasets-of-Sintel,
  title = {A naturalistic open source movie for optical flow evaluation},
  author = {Butler, D. J. and Wulff, J. and Stanley, G. B. and Black, M. J.},
  booktitle = {European Conf. on Computer Vision (ECCV)},
  editor = {{A. Fitzgibbon et al. (Eds.)}},
  publisher = {Springer-Verlag},
  series = {Part IV, LNCS 7577},
  month = oct,
  pages = {611--625},
  year = {2012}
}

@inproceedings{datasets-of-Spring,
  title={Spring: A high-resolution high-detail dataset and benchmark for scene flow, optical flow and stereo},
  author={Mehl, Lukas and Schmalfuss, Jenny and Jahedi, Azin and Nalivayko, Yaroslava and Bruhn, Andr{\'e}s},
  booktitle={Proceedings of the IEEE/CVF Conference on Computer Vision and Pattern Recognition},
  pages={4981--4991},
  year={2023}
}

@InProceedings{datasets-of-FlyingChiars,
  author    = "A. Dosovitskiy and P. Fischer and E. Ilg and P. H{\"a}usser and C. Haz{\i}rba{\c{s}} and V. Golkov and P. v.d. Smagt and D. Cremers and T. Brox",
  title     = "FlowNet: Learning Optical Flow with Convolutional Networks",
  booktitle = "IEEE International Conference on Computer Vision (ICCV)",
  month     = " ",
  year      = "2015",
  url       = "http://lmb.informatik.uni-freiburg.de/Publications/2015/DFIB15"
}

@InProceedings{datasets-of-FlyingChiars2,
  author    = "E. Ilg and T. Saikia and M. Keuper and T. Brox",
  title     = "Occlusions, Motion and Depth Boundaries with a Generic Network for Disparity, Optical Flow or Scene Flow Estimation",
  booktitle = "European Conference on Computer Vision (ECCV)",
  month     = " ",
  year      = "2018",
  url       = "http://lmb.informatik.uni-freiburg.de/Publications/2018/ISKB18"
}

\end{document}